\documentclass[conference]{IEEEtran}
\usepackage{times}

\usepackage[numbers]{natbib}
\usepackage{multicol}
\usepackage[bookmarks=true]{hyperref}
\usepackage{color}
\let\labelindent\relax
\usepackage{enumitem}
\usepackage{graphicx}
\usepackage{amsmath}
\usepackage{amsfonts}       
\usepackage{
tikz,
relsize,
booktabs
}
\usepackage[font={small}]{caption}
\usepackage{bbm}
\usepackage{tablefootnote}

\usepackage{colortbl}
\usepackage{algpseudocode}

\usepackage[normalem]{ulem}
\useunder{\uline}{\ul}{}

\usepackage{tikz}

\usepackage[ruled,vlined]{algorithm2e}

\definecolor{commentcolor}{RGB}{110,154,155}   
\newcommand{\PyComment}[1]{\ttfamily\textcolor{commentcolor}{\# #1}}  
\newcommand{\PyCode}[1]{\ttfamily\textcolor{black}{#1}} 

\newcommand{\mysection}[1]{%
  \vspace{0.25\baselineskip}
  \par\noindent\textbf{#1}:%
}

\usepackage{pifont}
\definecolor{olivegreen}{HTML}{3C8031}
\definecolor{tableblue}{HTML}{2D2F92}

\newcommand{\xmark}{\ding{55}}%
\newcommand{\redcross}{\textcolor{red}{\xmark}}

\newcommand{\cmark}{\ding{51}}%
\newcommand{\greencheck}{\textcolor{olivegreen}{\cmark}}

\renewcommand\thefigure{\arabic{figure}}
\newcommand{\xxnote}[3]{}
\ifx\hidenotes\undefined
  \renewcommand{\xxnote}[3]{}
\fi

\hypersetup{
    colorlinks=true,
    linkcolor=blue,
    filecolor=magenta,      
    citecolor=blue,
    urlcolor=teal,
}

\newcommand{\method}{\textsc{Open Teach}}
\newcommand{\website}{\url{https://open-teach.github.io/}}

\newcommand\superequiv{\mathrel{\rlap{\raisebox{\fontdimen22\textfont2}{$=$}}\raisebox{-0.5\fontdimen22\textfont2}{$ = $}}}

\IEEEoverridecommandlockouts

\begin{document}

\title{\method{}: A Versatile Teleoperation \\System for Robotic Manipulation}


\author{
Aadhithya Iyer$^{1}$\thanks{Correspondence to: aadhithya.iyer@nyu.edu} \qquad Zhuoran Peng$^{1}$ \qquad Yinlong Dai$^{1}$ \qquad Irmak Güzey$^{1}$ 
\\ 
Siddhant Haldar$^{1}$ \qquad Soumith Chintala$^{2}$ \qquad Lerrel Pinto$^{1}$ \\
\\
$^{1}$New York University \qquad $^{2}$Meta \\
\\
{\small \tt \website{}}
\vspace{-0.2in}
}

\makeatletter
\let\@oldmaketitle\@maketitle%
\renewcommand{\@maketitle}{\@oldmaketitle%
    \centering
    \vspace{0.3in}
    \includegraphics[width=\textwidth]{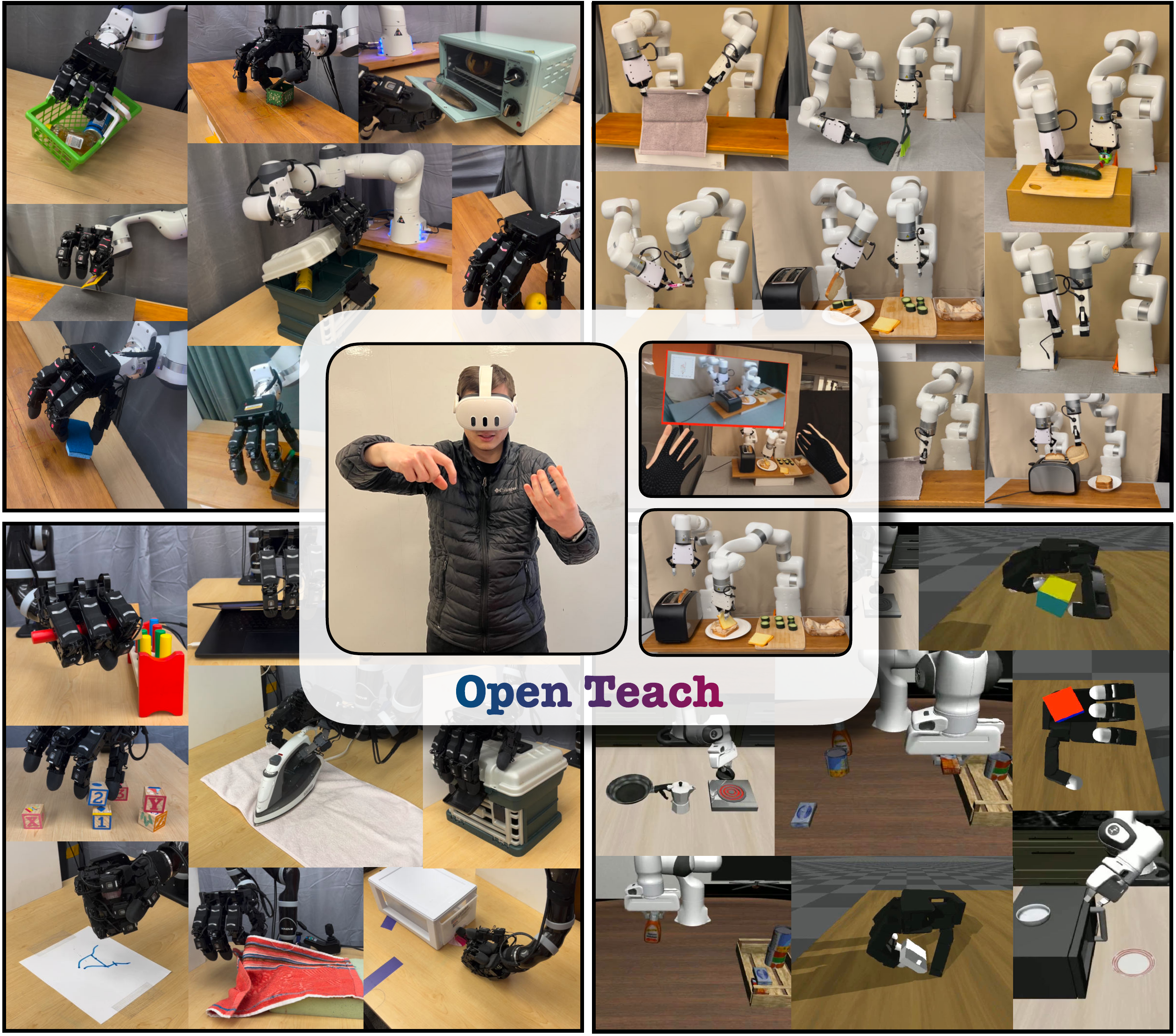}
    \captionof{figure}{We present \method{}, a unified robot teleoperation framework that supports multiple arms and hands, allows mobile manipulation, is calibration-free, and works across both simulation and real-world environments. \method{} uses a VR headset for teleoperation and offers low latency and high-frequency visual feedback. This high-frequency operation allows human users to correct for robot errors in real time, facilitating the execution of intricate, long-horizon tasks. From \textit{making a sandwich} and \textit{ironing cloth} to \textit{placing items in a basket and lifting it} and \textit{approaching a cabinet and opening it}, \method{} delivers a comprehensive, user-friendly teleoperation experience for a wide range of applications. \method{} is fully open-source.} 
    \label{fig:intro}
}
\makeatother

\maketitle

\begin{abstract}


Open-sourced, user-friendly tools form the bedrock of scientific advancement across disciplines. The widespread adoption of data-driven learning has led to remarkable progress in multi-fingered dexterity, bimanual manipulation, and applications ranging from logistics to home robotics. However, existing data collection platforms are often proprietary, costly, or tailored to specific robotic morphologies. We present \method{}, a new teleoperation system leveraging VR headsets to immerse users in mixed reality for intuitive robot control. Built on the affordable Meta Quest 3, which costs \$500, \method{} enables real-time control of various robots, including multi-fingered hands, bimanual arms, and mobile manipulators, through an easy-to-use app. Using natural hand gestures and movements, users can manipulate robots at up to 90Hz with smooth visual feedback and interface widgets offering closeup environment views. We demonstrate the versatility of \method{} across 38 tasks on different robots. A comprehensive user study indicates significant improvement in teleoperation capability over the AnyTeleop framework. Further experiments exhibit that the collected data is compatible with policy learning on 10 dexterous and contact-rich manipulation tasks. Currently supporting Franka, xArm, Jaco, Allegro, and Hello Stretch platforms, \method{} is fully open-sourced to promote broader adoption. Videos are available at \website{}.

\end{abstract}

\IEEEpeerreviewmaketitle

\section{Introduction}
\label{intro}

The integration of learning-based methods has sparked a revolution in robotics, significantly enhancing capabilities in manipulation~\cite{chi2022iterative, zhao2023learning, brohan2022rt, dobbe}, locomotion~\cite{gangapurwala2022rloc, ma2022combining, smith2022walk, cheng2023extreme}, and aerial robotics~\cite{zhang2016learning, gandhi2017learning, hwangbo2017control}.
More recent work has been making advancements in complex single-task behavior learning~\cite{wang2023mimicplay, arunachalam2023holo, zhao2023learning}, multitask scenarios~\cite{padalkar2023open,bharadhwaj2023roboagent}, multimodal behavior learning~\cite{shafiullah2022behavior, cui2022play, chi2023diffusion, reuss2023goal}, and efficient fine-tuning of pretrained behavior models~\cite{haldar2022watch, haldar2023teach, nair2020awac}. A fundamental requirement across all these threads of research is the need to collect data in the form of task demonstrations.


Commonly used teleoperation systems include devices such as joysticks and 3D spacemouses~\cite{liu2023libero, sian2004whole}, commercial VR headsets~\cite{gharaybeh2019telerobotic, zhang2018deep, arunachalam2022dexterous, arunachalam2023holo, MVP, george2023openvr}, kinesthetic teaching~\cite{billard2006discriminative}, and phone teleoperation~\cite{mandlekar2018roboturk}. The aforementioned devices are cost-effective and easy to set up. However, they are often unintuitive to use and require extensive user-training to demonstrate intricate motions.
Recently proposed exoskeleton-based teleoperation frameworks like ALOHA~\cite{zhao2023learning}, GELLO~\cite{wu2023gello}, and AirExo~\cite{fang2023low} attempt to alleviate this problem by having the human teleoperator directly control a kinematically isomorphic version of the same robot arm. These frameworks directly impose the kinematic constraints of the robot arm during teleoperation making it more compatible and intuitive to control the motion of the robot. Although highly effective, these systems can require an additional robot for each robot being controlled, have high initial setup costs, and are designed for specific robot morphologies.


The challenge of easy-to-use teleoperation devices is more apparent in dexterous manipulation problems~\cite{DexPilot, qin2023anyteleop, arunachalam2022dexterous, arunachalam2023holo}, owing to the high dimensional action space. Such frameworks typically involve the use of expensive gloves~\cite{glovereview, HAPTIX, li2020mobile}, extensive calibration processes~\cite{DexPilot, arunachalam2022dexterous}, or are susceptible to monocular occlusions~\cite{arunachalam2022dexterous}. 

In this work, we present \method{}, an open-source framework for robot teleoperation that supports a variety of robots, including bimanual and multi-finger manipulation, all at a price of \$500. As shown in Figure~\ref{fig:intro}, \method{} uses a VR headset (e.g. Quest 3) to put users / teachers in an immersive virtual world where they can view a robotic scene both through their eyes, via visual passthrough, as well as realtime streams from the robot's cameras. To control the robot, users can simply use hand gestures, which are detected using onboard hand-pose estimators at 90Hz. As a result, even though \method{} is kinematics-unaware, the high frequency execution and improved hand pose detection accuracy enables users to collect high-quality real-time robot demonstrations. 



We experimentally evaluate \method{} on 38 tasks across single arm, bimanual, multi-fingered, and mobile manipulation robot setups in both simulation and the real world. The tasks span from tabletop manipulation to contact-rich dexterous manipulation. Across different robot morphologies, we find that users can provide demonstrations at speeds on par with robot-specific teleoperation systems and significantly faster than general-purpose systems like AnyTeleop~\cite{qin2023anyteleop}.
Importantly, policies trained on the data collected achieve an average success rate of 86\% on 10 tasks in simulation and the real world, validating the utility of policy learning using \method{}. The contributions of this work is summarized as follows:



\begin{enumerate}
    \item We present \method{}, an open-source system for plug-and-play teleoperation framework suitable for collecting demonstrations across different robot morphologies in both simulation and the real world.
    \item We experimentally show that the demonstrations collected by \method{} can be used to train effective, general-purpose manipulation behaviors.
    \item Our user study on 15 users highlights the efficacy of \method{} for both experienced and new users.
\end{enumerate}

\method{} is fully open-source with the mixed reality API, policy training code, and demonstrations collected using \method{} available at \website{}.

\begin{table*}[!h]
\centering
\caption{Comparison of \method{}'s capabilities with prior teleoperation systems on features such as being calibration-free, compatible with multi-fingered hands, bimanual arms, and mobile manipulators, and being open-sourced.} 
\label{table:comparisons}
\small
\begin{tabular}{lcccccc}
\hline
                                                & Calibration Free     & Hands       & Arms           & Bimanual    & Mobile Manipulation & Open-source \\ \hline
Joystick                                        & \greencheck          & \redcross   & \greencheck    & \redcross   & \redcross           & \greencheck \\
Spacemouse                                      & \greencheck          & \redcross   & \greencheck    & \redcross   & \redcross           & \greencheck \\
Phone Teloperation~\cite{mandlekar2018roboturk} & \greencheck          & \redcross   & \greencheck    & \redcross   & \redcross           & \redcross   \\
DexPilot~\cite{DexPilot}                        & \redcross            & \greencheck & \greencheck    & \redcross   & \redcross           & \redcross   \\
Holo-Dex~\cite{arunachalam2023holo}             & \greencheck          & \greencheck & \redcross      & \redcross   & \redcross           & \greencheck \\
DIME~\cite{arunachalam2022dexterous}            & \redcross            & \greencheck & \redcross      & \redcross   & \redcross           & \greencheck \\
TeachNet~\cite{li2019vision}                    & \greencheck          & \greencheck & \redcross      & \redcross   & \redcross           & \greencheck \\
Telekinesis~\cite{sivakumar2022robotic}         & \greencheck          & \greencheck & \greencheck    & \redcross   & \redcross           & \redcross   \\
Qin et al.~\cite{qin2022one}                    & \greencheck          & \greencheck & \redcross      & \greencheck & \redcross           & \greencheck \\
MVP-Real~\cite{MVP}                             & \redcross            & \greencheck & \greencheck    & \redcross   & \redcross           & \redcross   \\
Transteleop~\cite{li2022dexterous}              & \redcross            & \greencheck & \greencheck    & \redcross   & \redcross           & \redcross   \\
Mosbach et al.~\cite{mosbach2022accelerating}   & \redcross            & \greencheck & \greencheck    & \redcross   & \redcross           & \greencheck \\
AnyTeleop~\cite{qin2023anyteleop}               & \greencheck          & \greencheck & \greencheck    & \redcross   & \redcross           & \redcross   \\
ALOHA~\cite{zhao2023learning}                   & \greencheck          & \redcross   & \greencheck    & \greencheck & \redcross           & \greencheck \\
Mobile ALOHA~\cite{fu2024mobile}                & \greencheck          & \redcross   & \greencheck    & \greencheck & \greencheck         & \greencheck \\
GELLO~\cite{wu2023gello}                        & \greencheck          & \redcross   & \greencheck    & \greencheck & \redcross           & \greencheck \\
AirExo~\cite{fang2023low}                       & \greencheck          & \redcross   & \greencheck    & \greencheck & \redcross           & \greencheck \\
Dobb-E~\cite{dobbe}                             & \greencheck          & \redcross   & \greencheck    & \redcross   & \greencheck         & \greencheck \\
\textbf{\textcolor{olivegreen}{\method{}}}      & \greencheck          & \greencheck & \greencheck    & \greencheck & \greencheck         & \greencheck \\ \hline
\end{tabular}
\end{table*}

\section{Related Work}
\label{related_work}

\subsection{Robot-Specific Teleoperation}
Teleoperation, as a medium for human-robot interaction, has been a crucial part of robotics. Recent strides in learning-based methods demand extensive data collection, giving rise to diverse teleoperation systems — joysticks and spacemouses~\cite{liu2023libero, sian2004whole}, VR controllers~\cite{gharaybeh2019telerobotic, zhang2018deep, arunachalam2022dexterous, arunachalam2023holo, MVP, george2023openvr}, RGB cameras~\cite{DexPilot,sivakumar2022robotic,qin2023anyteleop,song2020grasping}, IMU sensors~\cite{laghi2018shared,kim2009walking,wu2019teleoperation}, kinesthetic teaching~\cite{billard2006discriminative}, phone teleoperation~\cite{mandlekar2018roboturk}, gloves~\cite{glovereview, HAPTIX, li2020mobile}, marker-based motion capture systems~\cite{zhao2012combining,liu2021semi}, and reacher-grabber sticks~\cite{pari2021surprising,dobbe}. However, several challenges remain in the robot-specific nature of these devices. For instance, devices like joysticks, spacemouses, and phones are limited to controlling robot arms due to their lack of fidelity for multi-fingered hands. There have been systems developed to map the human hand pose to the robot pose~\cite{arunachalam2022dexterous,arunachalam2023holo, qin2023anyteleop,li2019vision,antotsiou2018task,meeker2020continuous}, but they are often restricted to only controlling robot hands. Further, all of these frameworks lack awareness of the robot's kinematic constraints, leading to challenges in intuitive control, especially in complex poses. As a solution, there are more conventional but expensive exo-skeleton based teleoperation systems~\cite{hulin2011dlr,katz2018low,schwarznimbro} that use a second arm for controlling the manipulator arm. Recently proposed ALOHA~\cite{zhao2023learning, fu2024mobile} affirms the effectiveness of this approach through impressive results in fine-grained bimanual manipulation. However, these systems require an exact copy of the manipulator robot arm, rendering them costly and less practical for heavier robots. In addressing these issues, GELLO~\cite{wu2023gello} and AirExo~\cite{fang2023low} introduce exo-skeleton teleoperation frameworks, utilizing a kinematically isomorphic variant of the robot arm. This approach proves more affordable and lightweight, enhancing usability for humans. Despite their success in fine-grained manipulation tasks, these solutions are constrained to robot arms and face challenges in extending seamlessly to control robot hands. For multi-fingered robot hands, gloves, vision-based, and VR-based methods have been employed. These systems either assume a fixed robot arm~\cite{arunachalam2022dexterous,arunachalam2023holo} or are tied to a specific robot setup~\cite{DexPilot,li2022dexterous}, making them difficult to transfer to new arm-hand systems and new environments.

\subsection{Unified Teleoperation Frameworks}
The robotics community has often sought to develop versatile systems that operate across diverse environments and robots~\cite{shah2023vint,padalkar2023open,octo_2023}. Leveraging the success of learning-based methods, achieving this requires teleoperation systems adaptable to various robot variants, allowing for abundant data collection with minimal setup costs. While methods combining robot arms with multi-fingered hands exist~\cite{DexPilot,sivakumar2022robotic,MVP,mosbach2022accelerating,li2022dexterous}, their applicability across robot variants remains unclear.  AnyTeleop~\cite{qin2023anyteleop} makes progress in this direction by proposing a robot-agnostic system compatible with multiple hands and arms. We build upon this idea in creating \method{}.

\setcounter{figure}{1} 
\begin{figure*}[!t]
    \centering
    \includegraphics[width=\textwidth]{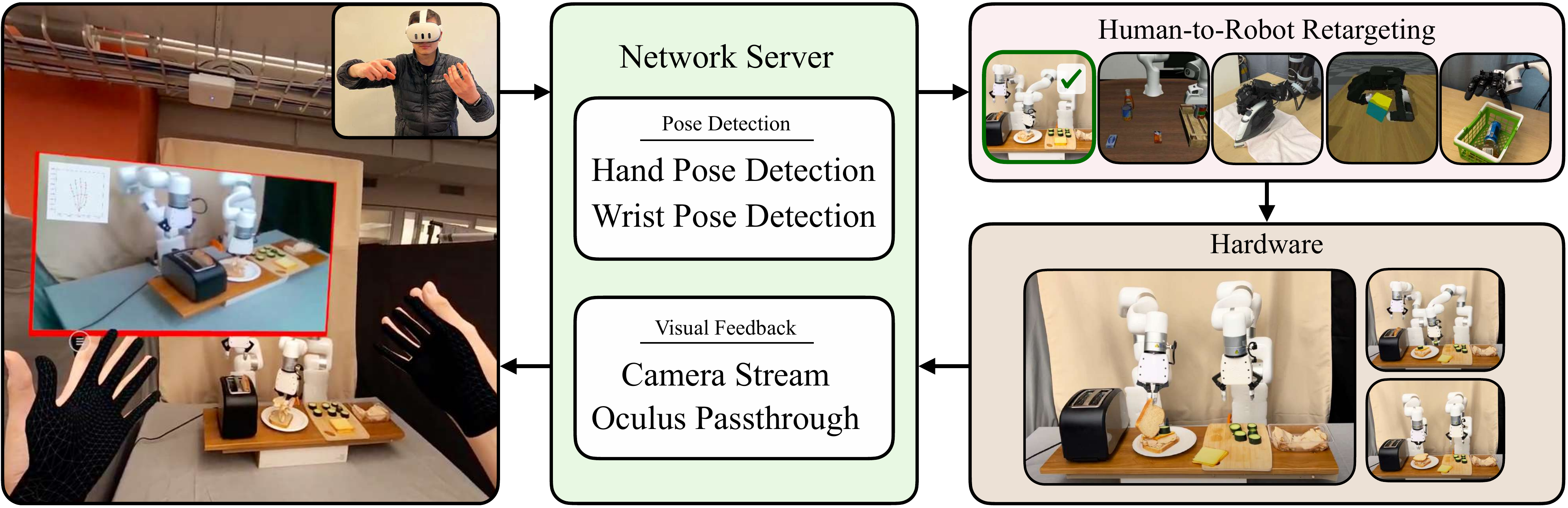}
    \caption{Overview of the teleoperation module in \method{}. Provided a hand and wrist pose within the VR interface, the controller transmits keypoint data to the robot's server. The server then transforms and retargets these key points to align with the specific robot setup. Real-time visual feedback of the teleoperated scene is promptly relayed back to the VR headset.}
    \label{fig:schematic}
\end{figure*}

\section{Background on Imitation Learning}
\label{background}

\subsection{Behavior Cloning}
Given a dataset of expert rollouts for a desired task in the form of observation and action pairs $\mathcal{D} \superequiv \{(o,a)\} \subset \mathcal{O}\times\mathcal{A}$, behavior cloning (BC) aims to learn a policy $\pi \colon \mathcal{O} \rightarrow \mathcal{A}$ that models this data without any online interactions with the environment nor a reward function. Often, such policies are chosen from a hypothesis class parameterized by a parameter set $\theta$. Following this convention, the objective of BC is to find the value $\theta$ that maximizes the probability of the observed data.

\begin{equation}
    \theta^{*} = \operatorname*{arg max}_{\theta} \prod_{t} \mathbb{P}(a_t | o_t; \theta)
\end{equation}

When constrained to unimodal isotropic  Gaussians, this maximum likelihood
estimation problem leads to minimizing the Mean Squared Error (MSE), $\Sigma_{t} \lVert a_t-\pi(o_t;\theta) \rVert^2$. 


\subsection{Inverse Reinforcement Learning}
In this work, we employ FISH~\cite{haldar2023teach} and TAVI~\cite{guzey2023see} to learn visual and visuotactile policies respectively. For both of these methods, the first phase involves obtaining a non-parametric base-policy $\pi^b \colon \mathcal{Z} \rightarrow \mathcal{A}$ with encoded representations $z \in \mathcal{Z}$ and actions $a \in \mathcal{A}$. Then a residual policy $\pi^r: \mathcal{Z}\times\mathcal{A}\rightarrow\mathcal{A}$ is learned atop the base policy $\pi^b$ such that an action sampled from the final policy $\pi$ is the sum of the base action $a^b\sim\pi^{b}(z)$ and the residual offset $a^r\sim\pi^{r}(z, a^b)$. The reward for learning the residual policy through inverse RL is obtained through optimal transport based trajectory matching~\cite{haldar2022watch, cohen2021imitation}. 

\section{\method{}}
\label{approach}

In \method{}, a user wears a Virtual Reality (VR) headset to provide demonstrations to a robot. This involves creating a virtual world for teaching, retargeting the teacher's hand and wrist pose to the robot joints, and finally controlling the robot. Table~\ref{table:comparisons} compares \method{} with various other teleoperation systems across a variety of robot types. We observe that \method{} is the only framework that enables controlling multiple arms, hands, and mobile manipulators, is calibration-free, and is completely open-source.

In this section, we provide details about the VR-based teleoperation setup and the system design that enables data collection using this framework.

\subsection{Placing an Operator in a Virtual World}
\begin{figure*}[!th]
    \centering
    \includegraphics[width=\textwidth]{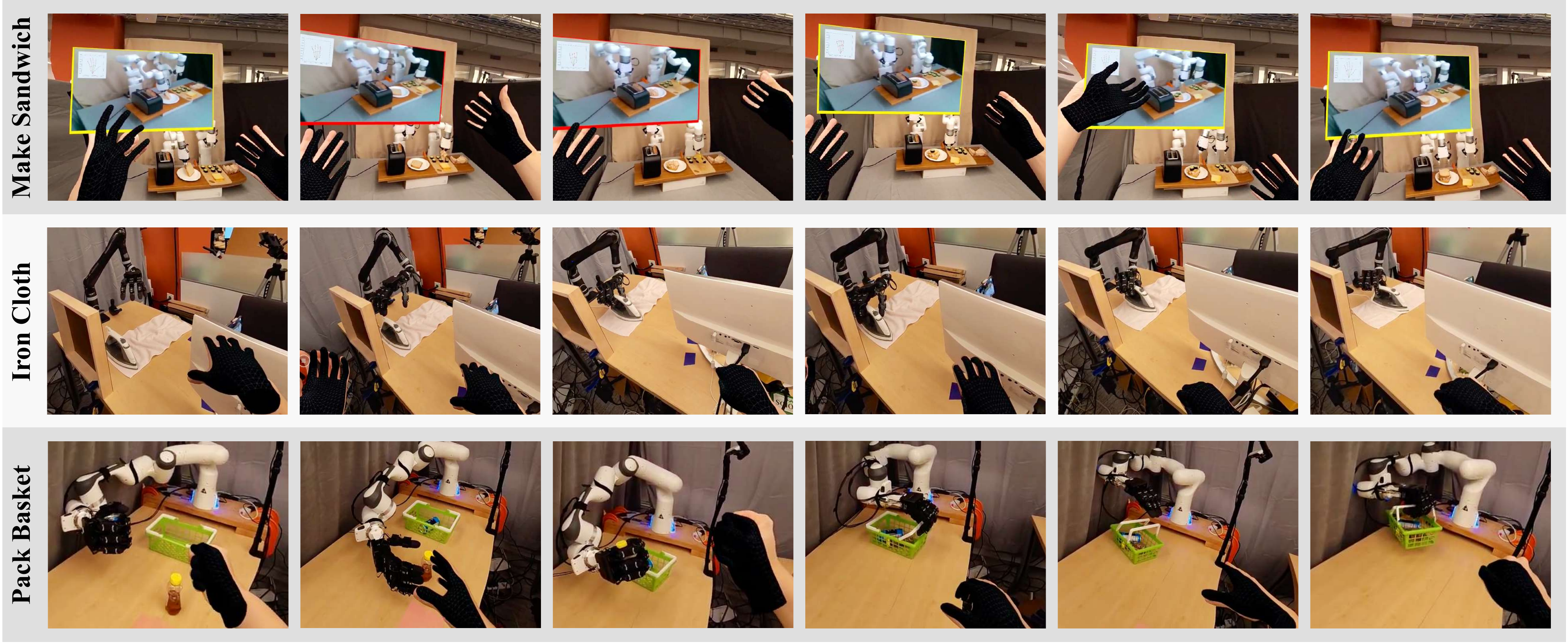}
    \caption{The demonstration collection process as viewed from within the VR application. Shown here is one task being performed for each real-world setup. High resolution images are streamed at 90 Hz to the VR application, allowing for an immersive experience and improved teleoperation compared to prior VR-based systems. The high frame rate streaming enables reactive control by the user, while widgets for visualizing the robot's camera view help the user focus on fine-grained movements.}
    \label{fig:demo_collection}
\end{figure*}

We use the Meta Quest 3 VR headset to place the human teacher in a virtual world. The headset surrounds the human in a virtual environment at a resolution of $2064\times2208$ and a native refresh rate of 90Hz. The base version of this headset is affordable at \$499 and is relatively light at 513g. Compared to the Meta Quest 2 VR headset used in prior work~\cite{arunachalam2023holo}, the Quest 3 provides a full-color passthrough allowing the human to get a direct view of the robot setup during teleoperation. These features, especially the full-color passthrough, allow for a comfortable and intuitive operation by the user. Additionally, similar to \citet{arunachalam2023holo}, the Quest 3 API interface allows for creating custom mixed reality worlds that visualize the robotic system along with diagnostic panels in VR. Examples of virtual scenes have been shown in Fig.~\ref{fig:schematic} and Fig.~\ref{fig:demo_collection}. It is important to highlight the exceptional clarity of the scene passthrough visible in Quest 3. The teacher can experience a 3D view of the scene through the headset, significantly enhancing the intuitiveness of the teleoperation experience.

\subsection{Pose Estimation with VR Headsets}
Similar to ~\citet{arunachalam2023holo}, we directly use the in-built hand pose estimator~\cite{megatrack} of the Quest 3 using 2 monochrome cameras. This is significantly more robust compared to single camera alternatives~\cite{zhang2020mediapipe}. Further, since the cameras are internally calibrated, they do not require specialized calibration routines that are needed in prior multi-camera teleoperation frameworks~\cite{DexPilot, qin2023anyteleop}. Also, since the hand-pose estimator is integrated into the device, it can stream real-time hand poses at 90Hz. This alleviates the challenge of obtaining hand poses at both high accuracy and high frequency, as reported in prior work~\cite{DexPilot, arunachalam2022dexterous}.

\subsection{Human to Robot Pose Retargeting}
\label{subsec:retargeting}
The inbuilt hand pose estimate from the VR headset provides us with the joint positions of all the fingers of the human hand and the wrist. With this information, we can design wrappers that use combinations of these joint positions to map the human hand poses to the robot poses for any given robot morphology. In this work, we use a variety of robot arms, each with either a 2-fingered gripper or a multi-fingered robot hand. Below, we provide the wrapper design for each robot morphology used in this work.

\mysection{Robot Arm} We use the wrist keypoint and the knuckle points of the index and pinky fingers to establish a 3D coordinate system with a 2D plane along the palm of the human hand, and a third axis perpendicular to the palm. Then, the wrist position is mapped to the robot end effector position, and the transformations of the 3D coordinate system across time are mapped to the changes in the end effector orientation.

\mysection{Robot Hand} We use the teacher's hand pose obtained from the VR to compute the individual joint angles in the teacher's hand. Given these joint angles, a straightforward method of retargeting is to directly command the robot’s joints to the corresponding angles. In practice, this works well for all fingers except the thumb. To address this, we improve upon \citet{arunachalam2023holo}, where the spatial coordinate of the teacher's thumb tip is mapped to that of the robot hand and then an inverse kinematics solver is used to compute the joint angles of the thumb. More details about the improvement in thumb retargeting have been included in Appendix~\ref{appendix:subsec:framework}. Since the Allegro hand does not have a pinky finger, we ignore the teacher’s pinky joints.

\mysection{Two-fingered gripper} To detect the opening and closing of the two-fingered gripper, we utilize the pinch between the pinky finger and the thumb. The pinch is detected by computing the distance between the tips of the two fingers and setting a threshold on the pinch distance. We use a toggle mechanism for opening and closing the gripper where each pinch indicates toggling to the alternate state of the gripper.

\mysection{Mobile manipulator} The same 3D coordinate system established for controlling robot arms is used for mapping the wrist's movements to actions of the mobile robot. When the wrist moves forward, it extends the robot's arm, enabling it to reach farther. Vertical wrist movements adjust the robot's height, while lateral wrist movements cause the robot to move sideways by controlling its wheels. The 3D transformations across time are mapped to the changes in the end effector orientation. The opening and closing of the gripper are controlled through the pinch between the index finger and the thumb.

These are just a few examples of how the hand pose data can be used to obtain a mapping between the human hand and a robot. The simplicity of the proposed framework has been summarized in Code Snippet~\ref{algo:hand_pose_retarget}. The primary idea behind \method{} is that given a VR headset, the end-user has the flexibility to design their own human-to-robot wrappers using the human hand poses as input. The framework has been designed for simple integration with any robot setup, allowing robot teleoperation with real-time streaming (up to 90Hz) and low-latency visual feedback. This significantly reduces the initial setup cost as compared to prior exoskeleton-based teleoperation frameworks like GELLO~\cite{wu2023gello} and AirExo~\cite{fang2023low}. To increase support for more robots, we invite roboticists to add support for their robots on our common \method{} GitHub repository (see Section~\ref{sec:invitation}).

\begin{algorithm}[!t]
\SetAlgorithmName{Code Snippet}{}
\SetAlgoLined
    \PyComment{Initialize robot and VR}\\
    \PyCode{vr = VR()} \\
    \PyCode{robot = Robot()} \\
    \PyCode{while(True):}\\
    \Indp
        \PyComment{Step 1: Get hand pose from VR} \\
        \PyCode{hPose = vr.getHandPose()}\\
        \PyComment{Step 2: Retarget to robot pose} \\
        \PyCode{rPose = robot.retargetH2R(hPose)}\\
        \PyComment{Step 3: Move robot to target pose} \\
        \PyCode{robot.move(rPose)} \\
    \Indm
\caption{Robot control using VR}
\label{algo:hand_pose_retarget}
\end{algorithm}

\subsection{Robot Control}
Achieving minimal error and low latency is pivotal for \method{} to facilitate the intuitive teleoperation of the robot hand by the human teacher. In this study, we employ the Allegro Hand as our robotic hand, controlling it asynchronously through the ROS~\cite{ros} communication framework. Using the computed robot joint positions from the retargeting procedure, a PD controller outputs desired torques at a frequency of 300Hz. To mitigate steady-state error, we include a gravity compensation module to compute offset torques.


We use three different robot arms for our evaluations — xArm, Franka Emika Panda, and Kinova Jaco. We use different controllers for each. The xArm is directly controlled through the official xArm API~\cite{xArmPythonSDK}. For the Franka Emika Panda, we use the Deoxys controller~\cite{zhu2022viola}. For the Kinova Jaco, we use the controller open-sourced by~\citet{arunachalam2022dexterous}. The Allegro hand's streaming frequency is configured at 60 Hz, while the xArm, Franka Emika Panka, and Kinova Jaco arms are set to 90 Hz, 60 Hz, and 60 Hz, respectively. Such high frequency teleoperation allows the human teacher to see the robot move in real time and immediately correct execution errors in the robot. The Hello Stretch is controlled at 5Hz using the controller released by \citet{dobbe}. Further, we acknowledge the fact that the human hand possesses fewer degrees of freedom than the robot. In response, we introduce a pause functionality, allowing the teacher to momentarily halt teleoperation, reorient themselves, and resume the process. Furthermore, we implement a resolution adjustment feature, which provides a performance boost for high-precision tasks such as delicately picking up a tea sachet. Details about these implementations have been included in Appendix~\ref{appendix:subsec:framework}.

\section{Experimental Evaluation}
\label{experiments}

Our experiments and tasks are designed to answer the following questions:

\begin{enumerate}
    \item How versatile is \method{} across a range of robotics setups?
    \item How successful are policies trained with \method{}?
    \item Can \method{} be used for performing complex, long-horizon tasks?
    \item How intuitive is the system for new users?
\end{enumerate}

\subsection{Experimental Setup}
We evaluate the versatility of \method{} by using it to collect demonstrations on six different setups — four in the real world and two in simulation. Each setup is a combination of a variant of a robot arm with either an Allegro Hand or a 2-fingered gripper. The real-world setups include:

\begin{enumerate}
    \item \textbf{Franka-Allegro:} Comprising a Franka Arm with an Allegro Hand having the Xela tactile sensors. 
    \item \textbf{Kinova-Allegro:} Comprising a Kinova Jaco Arm with an Allegro Hand with the Xela tactile sensors.
    \item \textbf{Bimanual:} Comprising 2 xArm7 robot arms with 2-fingered grippers.
    \item \textbf{Stretch:} Comprising a Hello Stretch mobile manipulator with a 2-fingered gripper.
\end{enumerate}

The Franka-Allegro and Kinova-Allegro comprise a single Intel Realsense camera for data collection, whereas the Bimanual setup collects the data from 5 different cameras. The Stretch has an iPhone attached to the wrist for data collection, similar to \citet{dobbe}. The simulated environments include:

\begin{enumerate}
    \item \textbf{Allegro Sim:} Comprising a floating Allegro Hand capable of performing both static and dynamic tasks.
    \item \textbf{LIBERO Sim~\cite{liu2023libero}:} Comprising a Franka Arm with a 2-fingered gripper placed in varied scenes.
\end{enumerate}

We demonstrate the usefulness of the collected data by training visual and visuotactile policies using behavior cloning~\cite{ALVINN} and inverse RL~\cite{Ng2000, Abbeel2004}.

\begin{figure*}[!h]
    \centering
    \includegraphics[width=0.95\textwidth]{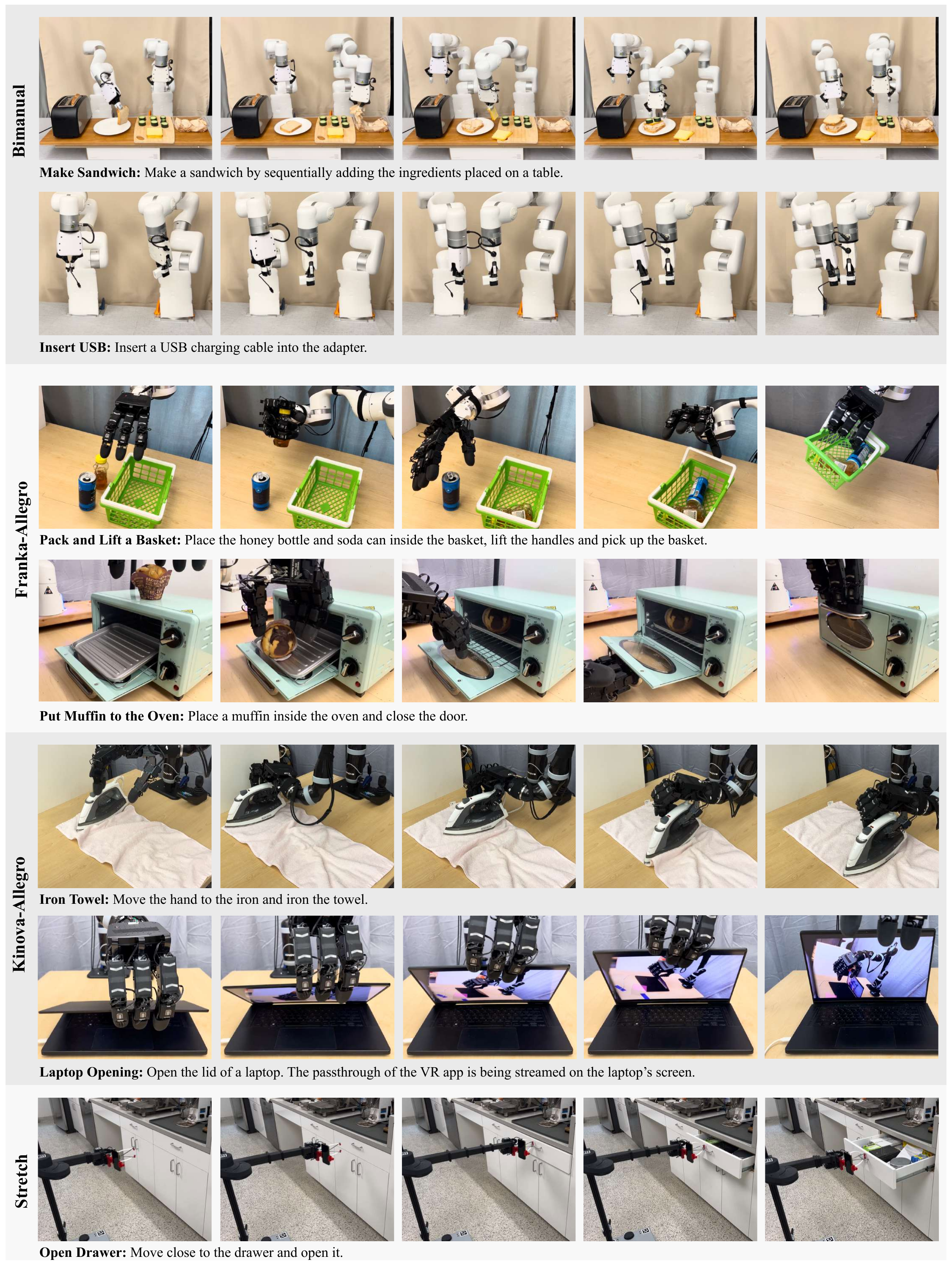}
    \caption{Real world task rollouts demonstrating the ability of \method{} to perform intricate, long-horizon tasks.}
    \label{fig:task_rollouts}
\end{figure*}

\subsection{Imitation Learning with \method{} Data}

Here, we describe the algorithms used for learning policies on data collected through \method{}.

\begin{enumerate}
    \item \textbf{Franka-Allegro:} We record both visual and tactile data for this setup. The policies are trained using TAVI~\cite{guzey2023see}, a demonstration-guided residual RL algorithm that collects a few expert demonstrations and learns a robot policy using both visual and tactile data.
    \item \textbf{Allegro Sim:} We only record visual data for this setup. The policies are trained using FISH~\cite{haldar2023teach}.
    \item \textbf{LIBERO Sim~\cite{liu2023libero}:} We only record visual data for this setup. The policies are trained using transformer-based BC with a GMM head~\cite{reynolds2009gaussian} and action chunking~\cite{zhao2023learning}.
\end{enumerate}

\begin{table*}[!th]
  \begin{minipage}{0.4\textwidth}
    \centering
    \caption{Teleoperation Frequency across all robots.}
    \label{table:robot_stats}
    \begin{tabular}{lccc}
    \hline
    \textbf{Domain} & \textbf{Robot Setup} & \multicolumn{2}{c}{\textbf{\begin{tabular}[c]{@{}c@{}}Stream \\ Frequency (in Hz)\end{tabular}}} \\ \cline{3-4} 
                    & \textbf{}            & \textbf{Arm}          & \textbf{End Effector}         \\ \hline 
    \rowcolor[HTML]{EFEFEF} 
    Real            & Franka-Allegro       & 60                    & 60                           \\
    \rowcolor[HTML]{EFEFEF} 
                    & Kinova-Allegro       & 60                    & 60                           \\
    \rowcolor[HTML]{EFEFEF} 
                    & Bimanual             & 90                    & 90                            \\
    \rowcolor[HTML]{EFEFEF} 
                    & Stretch             & 5                    & 5                            \\
    Sim      & Allegro Sim          & 60                    & 60                            \\
                    & LIBERO Sim           & 20                    & 20                            \\ \hline
    \end{tabular}
  \end{minipage}
  \begin{minipage}{0.6\textwidth}
    \centering
    \caption{Performance of policies learned on data collected through \method{}. For Franka-Allegro, Allegro Sim, and Libero Sim,  TAVI~\cite{guzey2023see}, FISH~\cite{haldar2023teach} and BC were respectively used to train policies. }
    \label{table:policy_performance}
    \begin{tabular}{lccc}
        \hline
        \rowcolor[HTML]{FFFFFF} 
        \textbf{Robot Setup}                   & \textbf{Task}                                                                     & \textbf{\begin{tabular}[c]{@{}c@{}}Number of\\ Demos\end{tabular}} & \textbf{\begin{tabular}[c]{@{}c@{}}Success\\ Rate\end{tabular}} \\ \hline
        \cellcolor[HTML]{EFEFEF}Franka-Allegro & \cellcolor[HTML]{EFEFEF}Open Box                                                  & \cellcolor[HTML]{EFEFEF}3                                          & \cellcolor[HTML]{EFEFEF}9/10                                                            \\
        \rowcolor[HTML]{EFEFEF} 
                                               & Grasp Sponge                                                                      & 6                                                                  & 7/10                                                            \\
        \cellcolor[HTML]{EFEFEF}               & \cellcolor[HTML]{EFEFEF}Pick Up Tea Sachet                                       & \cellcolor[HTML]{EFEFEF}4                                          & \cellcolor[HTML]{EFEFEF}7/10                                                            \\
        \rowcolor[HTML]{EFEFEF} 
                                               & Grasp Object and Twist                                                             & 6                                                                  & 8/10                                                            \\
        \rowcolor[HTML]{FFFFFF} 
        Allegro Sim                            & Flip Cube                                                                         & 6                                                                  & 10/10                                                           \\
        \rowcolor[HTML]{FFFFFF} 
                                               & Flip Sponge                                                                       & 6                                                                  & 10/10                                                           \\
        \rowcolor[HTML]{FFFFFF} 
                                               & Pinch Grasp                                                                       & 6                                                                  &  7/10                                                               \\
        \rowcolor[HTML]{EFEFEF} 
        Libero Sim                             & Close Top Drawer of Cabinet                                                       & 10                                                                 & 10/10                                                           \\
        \rowcolor[HTML]{EFEFEF} 
                                               & Turn on Stove                                                                     & 10                                                                 & 9/10                                                            \\
        \rowcolor[HTML]{EFEFEF} 
                                               & \begin{tabular}[c]{@{}c@{}}Pick and Place Soup \\ into Basket\end{tabular} & 50                                                                 & 9/10                                                            \\ \hline
  \end{tabular}
  \end{minipage}
\end{table*}

\begin{table*}[!htbp]
\centering
\caption{User study comparing \method{} with other baselines when used by experts and new users.}
\label{table:user_study}
\begin{tabular}{l|cccc|cccc}
\hline
\textbf{Task}     & \multicolumn{4}{c|}{\textbf{Success Rate}}                                                                      & \multicolumn{4}{c}{\textbf{\begin{tabular}[c]{@{}c@{}}Median completion time for\\ successful demonstrations (in s)\end{tabular}}} \\ \hline
\textbf{}         & \multicolumn{3}{c|}{\textbf{New User}}                                                    & \textbf{Expert}     & \multicolumn{3}{c|}{\textbf{New User}}                                                                    & \textbf{Expert}         \\ \cline{2-9} 
\textbf{}         & \textbf{Holo-Dex} & \textbf{AnyTeleop} & \multicolumn{1}{c|}{\textbf{Open Teach}}         & \textbf{Open Teach} & \textbf{Holo-Dex}      & \textbf{AnyTeleop}      & \multicolumn{1}{c|}{\textbf{Open Teach}}               & \textbf{Open Teach}     \\ \hline
\rowcolor[HTML]{EFEFEF} 
Flip cube         & 1                 & 1               & \multicolumn{1}{c|}{\cellcolor[HTML]{EFEFEF}1}      &  1                  & 6.58                   & 13.71                   & \multicolumn{1}{c|}{\cellcolor[HTML]{EFEFEF}5.5}      &  2.85                   \\
Pinch Grasp       & 0              & 0.2               & \multicolumn{1}{c|}{0.8}                             &  1                  & 17.49                  & 18.94                   & \multicolumn{1}{c|}{18.72}                             & 3.71                        \\
\rowcolor[HTML]{EFEFEF} 
Pour              & N/A               & N/A                & \multicolumn{1}{c|}{\cellcolor[HTML]{EFEFEF}0.4} &  0.8                & N/A                    & N/A                     & \multicolumn{1}{c|}{\cellcolor[HTML]{EFEFEF}40.97}     & 14.83                        \\
Pick and Place    & N/A               & N/A                & \multicolumn{1}{c|}{0.8}                         &  0.8                & N/A                    & N/A                     & \multicolumn{1}{c|}{23.57}                             & 11.875                        \\
\rowcolor[HTML]{EFEFEF} 
Open box of mints & N/A               & N/A                & \multicolumn{1}{c|}{\cellcolor[HTML]{EFEFEF}0.5} & 1                 & N/A                    & N/A                     & \multicolumn{1}{c|}{\cellcolor[HTML]{EFEFEF}32.21}     & 20.45                        \\ \hline
\end{tabular}
\end{table*}

\subsection{How versatile is \method{} across robotic setups?}
The primary idea behind \method{} is that given any robotic setup, a user can purchase an affordable off-the-shelf VR headset (in this case, Quest 3) and plug the headset and robot setup into the proposed framework to start teleoperating the robot without any additional hardware setup cost. To investigate its versatility, we use \method{} to teleoperate four different real world robotic setups, each having a different combination of a robot arm and end effector type — Franka Allegro, Kinova Allegro, a Bimanual setup with 2 xArm7 robots, and Hello Stretch for mobile manipulation. We also exhibit the applicability of \method{} in simulation through evaluations on 2 simulated environment suites — Allegro Sim and LIBERO Sim~\cite{liu2023libero}. The frequency of teleoperation for each of the setups has been provided in Table~\ref{table:robot_stats}. Table~\ref{table:policy_performance} provides a set of tasks performed on Franka-Allegro, Allegro Sim, and LIBERO Sim. A more comprehensive list of tasks, including those on the Kinova-Allegro, Bimanual, and Stretch setup have been provided in Fig.~\ref{fig:task_rollouts} and  Appendix~\ref{appendix:subsubsec:task_desc}. 

\subsection{How successful are policies trained with \method{}?}
Table~\ref{table:policy_performance} provides the success rates of policies learned using imitation learning across both the real-world and simulated setups. We use TAVI~\cite{guzey2023see} to learn visuotactile policies on Franka-Allegro, and FISH~\cite{haldar2023teach} to learn visual policies on Allegro Sim. Similar to prior work~\cite{guzey2023see, haldar2023teach}, these policies were learned within 20 minutes and achieved an average success rate of 82\%, validating the high quality of the observation data collected through \method{}. We employ behavior cloning to train policies on LIBERO Sim, achieving an average success rate of 93\%, thus confirming the high quality of the collected action data. Overall, the learned policies achieve an average success rate of 86\% across all tasks and robot morphologies. This underscores the effectiveness of \method{} in collecting data for policy learning.

\subsection{Can \method{} be used for performing complex, long-horizon tasks?}
In this section, we emphasize the efficacy of \method{} in executing a diverse array of complex, long-horizon tasks across various robotic configurations. Illustrated in Fig.~\ref{fig:task_rollouts} are examples of real-world task rollouts for the Bimanual, Franka-Allegro, Kinova-Allegro and Stretch setups. \method{} allows the collection of demonstrations for intricate, extended tasks, ranging from high-precision activities like USB insertion to delicate movements such as slicing a cucumber. On the multi-fingered hand setup, we demonstrate a broad spectrum of tasks, encompassing extended activities like \textit{placing objects in a basket and lifting it} to contact-rich manipulation scenarios like \textit{opening a laptop} and \textit{sliding a tea sachet off the table}. A detailed compilation of tasks performed in both real-world and simulated setups, along with more task rollouts, have been included in Appendix~\ref{appendix:subsec:robot_task_details}. Videos showcasing these task rollouts can be found on our project website.


\subsection{How intuitive is the system for new users?}
\label{subsec:user_study}
We assess the user-friendliness of the \method{} through a comprehensive user study involving 15 new users. The study is carried out on the Franka-Allegro setup, chosen for its capacity to evaluate the system's performance on both the robot hand and the robot arm. Each participant is allocated a 10-minute practice session to familiarize themselves with teleoperating the robot setup. Following this, they are tasked with performing five trials for each of three distinct tasks using Holo-Dex~\cite{arunachalam2023holo}, AnyTeleop~\cite{qin2023anyteleop}, and \method{}. To mitigate potential biases, the order of tasks is randomized for each user.

In Table~\ref{table:user_study}, we present a comparative analysis of success rates and median completion times for new users across Holo-Dex, AnyTeleop, and \method{} for the tasks of cube flipping and pinch grasping. We chose to analyze the median rather than mean completion times in order to mitigate potential biases from outliers, given the relatively small user sample size. Since the Holo-Dex and AnyTeleop baselines lack open-source code for arm retargeting, we were unable to evaluate them on tasks involving arm movements. Thus, our comparison is limited to the cube flipping and pinch grasping tasks that do not require arm manipulation. On these tasks, \method{} demonstrates a higher success rate along with significantly reduced median time to complete tasks compared to the other baselines.

Table~\ref{table:user_study} also includes a comparison of success rates and median completion times between an expert and new users utilizing \method{} for all tasks. On average, new users exhibit a success rate that is 76\% of the expert's and take $2.25\times$ longer to complete a task. Additional details regarding individual user performances are provided in Appendix~\ref{appendix:user_study}. Intriguingly, some new users, despite their unfamiliarity with the framework, achieve comparable or superior performance to the experts in certain tasks. This observation highlights two factors: (1) the inherent variation in abilities among individuals, and (2) while our system is intuitive for new users, prolonged training leads to substantial improvement in their performance, with the potential for further enhancement with continued practice.

\section{Limitations and Discussion}
\label{discussion}

In this work, we introduce \method{}, an open-source unified framework designed to facilitate low-latency, high-frequency robot teleoperation. This versatile framework is tailored to accommodate diverse tasks and is compatible with a range of robot morphologies. However, we recognize a few limitations in this work: $(a)$ \method{} relies on the accuracy of the in-built hand pose detection in the VR headset. Inaccuracies, particularly when fingers are occluded from view, can diminish the quality of hand tracking, posing challenges to teleoperation. $(b)$ In specific instances, the pose detector on the Oculus board may misconstrue finger positions, leading to difficulties in executing gestures like gripper closing, which relies on precise pinches between fingers. Addressing these challenges through future research on hand pose detection and tracking holds the potential to enhance the ease and intuitiveness of teleoperation using VR headsets.


\section{Invitation for Contributions to \method{}}
\label{sec:invitation}

We firmly advocate for the advancement of robotics through the unrestricted accessibility of research projects to the broader community. In line with this principle, we will open-source every component of this research endeavor, including the VR application, the human-robot interface, and the robot controllers. Further, we encourage fellow researchers in establishing the infrastructure for their own robotic setups, and we invite inquiries for assistance. Finally, we view our work as a step towards achieving more accessible and affordable robot teleoperation. Recognizing that there is ample room for enhancement, we enthusiastically welcome contributions to our repositories and would be happy to share such contributions with the world with proper attribution given to the contributors.



\section*{Acknowledgments}
We thank Mahi Shafiullah, Raunaq Bhirangi, Yibin Wang, Venkatesh Pattabiraman, Haritheja Etukuru and Chenyu Wang for valuable feedback and discussions. This work was supported by grants from Honda, Meta, Amazon, and ONR awards N00014-21-1-2758 and N00014-22-1-2773.

\bibliographystyle{plainnat}
\bibliography{references}

\begin{thebibliography}{69}
\providecommand{\natexlab}[1]{#1}
\providecommand{\url}[1]{\texttt{#1}}
\expandafter\ifx\csname urlstyle\endcsname\relax
  \providecommand{\doi}[1]{doi: #1}\else
  \providecommand{\doi}{doi: \begingroup \urlstyle{rm}\Url}\fi

\bibitem[Abbeel and Ng(2004)]{Abbeel2004}
Pieter Abbeel and Andrew~Y Ng.
\newblock Apprenticeship learning via inverse reinforcement learning.
\newblock In \emph{ICML}, 2004.

\bibitem[Antotsiou et~al.(2018)Antotsiou, Garcia-Hernando, and Kim]{antotsiou2018task}
Dafni Antotsiou, Guillermo Garcia-Hernando, and Tae-Kyun Kim.
\newblock Task-oriented hand motion retargeting for dexterous manipulation imitation.
\newblock In \emph{Proceedings of the European Conference on Computer Vision (ECCV) Workshops}, pages 0--0, 2018.

\bibitem[Arunachalam et~al.(2022)Arunachalam, Silwal, Evans, and Pinto]{arunachalam2022dexterous}
Sridhar~Pandian Arunachalam, Sneha Silwal, Ben Evans, and Lerrel Pinto.
\newblock Dexterous imitation made easy: A learning-based framework for efficient dexterous manipulation.
\newblock \emph{arXiv preprint arXiv:2203.13251}, 2022.

\bibitem[Arunachalam et~al.(2023)Arunachalam, G{\"u}zey, Chintala, and Pinto]{arunachalam2023holo}
Sridhar~Pandian Arunachalam, Irmak G{\"u}zey, Soumith Chintala, and Lerrel Pinto.
\newblock Holo-dex: Teaching dexterity with immersive mixed reality.
\newblock In \emph{2023 IEEE International Conference on Robotics and Automation (ICRA)}, pages 5962--5969. IEEE, 2023.

\bibitem[Bharadhwaj et~al.(2023)Bharadhwaj, Vakil, Sharma, Gupta, Tulsiani, and Kumar]{bharadhwaj2023roboagent}
Homanga Bharadhwaj, Jay Vakil, Mohit Sharma, Abhinav Gupta, Shubham Tulsiani, and Vikash Kumar.
\newblock Roboagent: Generalization and efficiency in robot manipulation via semantic augmentations and action chunking.
\newblock \emph{arXiv preprint arXiv:2309.01918}, 2023.

\bibitem[Billard et~al.(2006)Billard, Calinon, and Guenter]{billard2006discriminative}
Aude~G Billard, Sylvain Calinon, and Florent Guenter.
\newblock Discriminative and adaptive imitation in uni-manual and bi-manual tasks.
\newblock \emph{Robotics and Autonomous Systems}, 54\penalty0 (5):\penalty0 370--384, 2006.

\bibitem[Brohan et~al.(2022)Brohan, Brown, Carbajal, Chebotar, Dabis, Finn, Gopalakrishnan, Hausman, Herzog, Hsu, et~al.]{brohan2022rt}
Anthony Brohan, Noah Brown, Justice Carbajal, Yevgen Chebotar, Joseph Dabis, Chelsea Finn, Keerthana Gopalakrishnan, Karol Hausman, Alex Herzog, Jasmine Hsu, et~al.
\newblock Rt-1: Robotics transformer for real-world control at scale.
\newblock \emph{arXiv preprint arXiv:2212.06817}, 2022.

\bibitem[Caeiro-Rodríguez et~al.(2021)Caeiro-Rodríguez, Otero-González, Mikic-Fonte, and Llamas-Nistal]{glovereview}
Manuel Caeiro-Rodríguez, Iván Otero-González, Fernando~A. Mikic-Fonte, and Martín Llamas-Nistal.
\newblock A systematic review of commercial smart gloves: Current status and applications.
\newblock \emph{Sensors}, 2021.
\newblock ISSN 1424-8220.
\newblock \doi{10.3390/s21082667}.

\bibitem[Cheng et~al.(2023)Cheng, Shi, Agarwal, and Pathak]{cheng2023extreme}
Xuxin Cheng, Kexin Shi, Ananye Agarwal, and Deepak Pathak.
\newblock Extreme parkour with legged robots.
\newblock \emph{arXiv preprint arXiv:2309.14341}, 2023.

\bibitem[Chi et~al.(2022)Chi, Burchfiel, Cousineau, Feng, and Song]{chi2022iterative}
Cheng Chi, Benjamin Burchfiel, Eric Cousineau, Siyuan Feng, and Shuran Song.
\newblock Iterative residual policy: for goal-conditioned dynamic manipulation of deformable objects.
\newblock \emph{arXiv preprint arXiv:2203.00663}, 2022.

\bibitem[Chi et~al.(2023)Chi, Feng, Du, Xu, Cousineau, Burchfiel, and Song]{chi2023diffusion}
Cheng Chi, Siyuan Feng, Yilun Du, Zhenjia Xu, Eric Cousineau, Benjamin Burchfiel, and Shuran Song.
\newblock Diffusion policy: Visuomotor policy learning via action diffusion.
\newblock \emph{arXiv preprint arXiv:2303.04137}, 2023.

\bibitem[Cohen et~al.(2021)Cohen, Amos, Deisenroth, Henaff, Vinitsky, and Yarats]{cohen2021imitation}
Samuel Cohen, Brandon Amos, Marc~Peter Deisenroth, Mikael Henaff, Eugene Vinitsky, and Denis Yarats.
\newblock Imitation learning from pixel observations for continuous control.
\newblock 2021.

\bibitem[Cui et~al.(2022)Cui, Wang, Muhammad, Pinto, et~al.]{cui2022play}
Zichen~Jeff Cui, Yibin Wang, Nur Muhammad, Lerrel Pinto, et~al.
\newblock From play to policy: Conditional behavior generation from uncurated robot data.
\newblock \emph{arXiv preprint arXiv:2210.10047}, 2022.

\bibitem[Fang et~al.(2023)Fang, Fang, Wang, Ren, Chen, Zhang, Wang, and Lu]{fang2023low}
Hongjie Fang, Hao-Shu Fang, Yiming Wang, Jieji Ren, Jingjing Chen, Ruo Zhang, Weiming Wang, and Cewu Lu.
\newblock Low-cost exoskeletons for learning whole-arm manipulation in the wild.
\newblock \emph{arXiv preprint arXiv:2309.14975}, 2023.

\bibitem[Fu et~al.(2024)Fu, Zhao, and Finn]{fu2024mobile}
Zipeng Fu, Tony~Z Zhao, and Chelsea Finn.
\newblock Mobile aloha: Learning bimanual mobile manipulation with low-cost whole-body teleoperation.
\newblock \emph{arXiv preprint arXiv:2401.02117}, 2024.

\bibitem[Gandhi et~al.(2017)Gandhi, Pinto, and Gupta]{gandhi2017learning}
Dhiraj Gandhi, Lerrel Pinto, and Abhinav Gupta.
\newblock Learning to fly by crashing.
\newblock In \emph{2017 IEEE/RSJ International Conference on Intelligent Robots and Systems (IROS)}, pages 3948--3955. IEEE, 2017.

\bibitem[Gangapurwala et~al.(2022)Gangapurwala, Geisert, Orsolino, Fallon, and Havoutis]{gangapurwala2022rloc}
Siddhant Gangapurwala, Mathieu Geisert, Romeo Orsolino, Maurice Fallon, and Ioannis Havoutis.
\newblock Rloc: Terrain-aware legged locomotion using reinforcement learning and optimal control.
\newblock \emph{IEEE Transactions on Robotics}, 2022.

\bibitem[George et~al.(2023)George, Bartsch, and Farimani]{george2023openvr}
Abraham George, Alison Bartsch, and Amir~Barati Farimani.
\newblock Openvr: Teleoperation for manipulation.
\newblock \emph{arXiv preprint arXiv:2305.09765}, 2023.

\bibitem[Gharaybeh et~al.(2019)Gharaybeh, Chizeck, and Stewart]{gharaybeh2019telerobotic}
Zaid Gharaybeh, Howard Chizeck, and Andrew Stewart.
\newblock Telerobotic control in virtual reality.
\newblock In \emph{OCEANS 2019 MTS/IEEE SEATTLE}, pages 1--8, 2019.
\newblock \doi{10.23919/OCEANS40490.2019.8962616}.

\bibitem[Guzey et~al.(2023)Guzey, Dai, Evans, Chintala, and Pinto]{guzey2023see}
Irmak Guzey, Yinlong Dai, Ben Evans, Soumith Chintala, and Lerrel Pinto.
\newblock See to touch: Learning tactile dexterity through visual incentives.
\newblock \emph{arXiv preprint arXiv:2309.12300}, 2023.

\bibitem[Haldar et~al.(2022)Haldar, Mathur, Yarats, and Pinto]{haldar2022watch}
Siddhant Haldar, Vaibhav Mathur, Denis Yarats, and Lerrel Pinto.
\newblock Watch and match: Supercharging imitation with regularized optimal transport.
\newblock \emph{arXiv preprint arXiv:2206.15469}, 2022.

\bibitem[Haldar et~al.(2023)Haldar, Pari, Rai, and Pinto]{haldar2023teach}
Siddhant Haldar, Jyothish Pari, Anant Rai, and Lerrel Pinto.
\newblock Teach a robot to fish: Versatile imitation from one minute of demonstrations.
\newblock \emph{arXiv preprint arXiv:2303.01497}, 2023.

\bibitem[Han et~al.(2020)Han, Liu, Cabezas, Twigg, Zhang, Petkau, Yu, Tai, Akbay, Wang, Nitzan, Dong, Ye, Tao, Wan, and Wang]{megatrack}
Shangchen Han, Beibei Liu, Randi Cabezas, Christopher~D. Twigg, Peizhao Zhang, Jeff Petkau, Tsz-Ho Yu, Chun-Jung Tai, Muzaffer Akbay, Zheng Wang, Asaf Nitzan, Gang Dong, Yuting Ye, Lingling Tao, Chengde Wan, and Robert Wang.
\newblock Megatrack: Monochrome egocentric articulated hand-tracking for virtual reality.
\newblock 2020.

\bibitem[Handa et~al.(2020)Handa, Van~Wyk, Yang, Liang, Chao, Wan, Birchfield, Ratliff, and Fox]{DexPilot}
Ankur Handa, Karl Van~Wyk, Wei Yang, Jacky Liang, Yu-Wei Chao, Qian Wan, Stan Birchfield, Nathan Ratliff, and Dieter Fox.
\newblock Dexpilot: Vision-based teleoperation of dexterous robotic hand-arm system.
\newblock In \emph{2020 IEEE International Conference on Robotics and Automation (ICRA)}, pages 9164--9170, 2020.
\newblock \doi{10.1109/ICRA40945.2020.9197124}.

\bibitem[Hulin et~al.(2011)Hulin, Hertkorn, Kremer, Sch{\"a}tzle, Artigas, Sagardia, Zacharias, and Preusche]{hulin2011dlr}
Thomas Hulin, Katharina Hertkorn, Philipp Kremer, Simon Sch{\"a}tzle, Jordi Artigas, Mikel Sagardia, Franziska Zacharias, and Carsten Preusche.
\newblock The dlr bimanual haptic device with optimized workspace.
\newblock In \emph{2011 IEEE International Conference on Robotics and Automation}, pages 3441--3442. IEEE, 2011.

\bibitem[Hwangbo et~al.(2017)Hwangbo, Sa, Siegwart, and Hutter]{hwangbo2017control}
Jemin Hwangbo, Inkyu Sa, Roland Siegwart, and Marco Hutter.
\newblock Control of a quadrotor with reinforcement learning.
\newblock \emph{IEEE Robotics and Automation Letters}, 2\penalty0 (4):\penalty0 2096--2103, 2017.
\newblock \doi{10.1109/LRA.2017.2720851}.

\bibitem[Katz(2018)]{katz2018low}
Benjamin~G Katz.
\newblock \emph{A low cost modular actuator for dynamic robots}.
\newblock PhD thesis, Massachusetts Institute of Technology, 2018.

\bibitem[Kim et~al.(2009)Kim, Hong, and Kim]{kim2009walking}
Sung-Kyun Kim, Seokmin Hong, and Doik Kim.
\newblock A walking motion imitation framework of a humanoid robot by human walking recognition from imu motion data.
\newblock In \emph{2009 9th IEEE-RAS International Conference on Humanoid Robots}, pages 343--348. IEEE, 2009.

\bibitem[Kumar and Todorov(2015)]{HAPTIX}
Vikash Kumar and Emanuel Todorov.
\newblock Mujoco haptix: A virtual reality system for hand manipulation.
\newblock In \emph{2015 IEEE-RAS 15th International Conference on Humanoid Robots (Humanoids)}, pages 657--663, 2015.
\newblock \doi{10.1109/HUMANOIDS.2015.7363441}.

\bibitem[Laghi et~al.(2018)Laghi, Maimeri, Marchand, Leparoux, Catalano, Ajoudani, and Bicchi]{laghi2018shared}
Marco Laghi, Michele Maimeri, Mathieu Marchand, Clara Leparoux, Manuel Catalano, Arash Ajoudani, and Antonio Bicchi.
\newblock Shared-autonomy control for intuitive bimanual tele-manipulation.
\newblock In \emph{2018 IEEE-RAS 18th International Conference on Humanoid Robots (Humanoids)}, pages 1--9. IEEE, 2018.

\bibitem[Li et~al.(2019)Li, Ma, Liang, G{\"o}rner, Ruppel, Fang, Sun, and Zhang]{li2019vision}
Shuang Li, Xiaojian Ma, Hongzhuo Liang, Michael G{\"o}rner, Philipp Ruppel, Bin Fang, Fuchun Sun, and Jianwei Zhang.
\newblock Vision-based teleoperation of shadow dexterous hand using end-to-end deep neural network.
\newblock In \emph{2019 International Conference on Robotics and Automation (ICRA)}, pages 416--422. IEEE, 2019.

\bibitem[Li et~al.(2020)Li, Jiang, Ruppel, Liang, Ma, Hendrich, Sun, and Zhang]{li2020mobile}
Shuang Li, Jiaxi Jiang, Philipp Ruppel, Hongzhuo Liang, Xiaojian Ma, Norman Hendrich, Fuchun Sun, and Jianwei Zhang.
\newblock A mobile robot hand-arm teleoperation system by vision and imu.
\newblock In \emph{2020 IEEE/RSJ International Conference on Intelligent Robots and Systems (IROS)}, pages 10900--10906. IEEE, 2020.

\bibitem[Li et~al.(2022)Li, Hendrich, Liang, Ruppel, Zhang, and Zhang]{li2022dexterous}
Shuang Li, Norman Hendrich, Hongzhuo Liang, Philipp Ruppel, Changshui Zhang, and Jianwei Zhang.
\newblock A dexterous hand-arm teleoperation system based on hand pose estimation and active vision.
\newblock \emph{IEEE Transactions on Cybernetics}, 2022.

\bibitem[Liu et~al.(2023)Liu, Zhu, Gao, Feng, Liu, Zhu, and Stone]{liu2023libero}
Bo~Liu, Yifeng Zhu, Chongkai Gao, Yihao Feng, Qiang Liu, Yuke Zhu, and Peter Stone.
\newblock Libero: Benchmarking knowledge transfer for lifelong robot learning.
\newblock \emph{arXiv preprint arXiv:2306.03310}, 2023.

\bibitem[Liu et~al.(2021)Liu, Jiang, Xu, Liu, and Wang]{liu2021semi}
Shaowei Liu, Hanwen Jiang, Jiarui Xu, Sifei Liu, and Xiaolong Wang.
\newblock Semi-supervised 3d hand-object poses estimation with interactions in time.
\newblock In \emph{Proceedings of the IEEE/CVF Conference on Computer Vision and Pattern Recognition}, pages 14687--14697, 2021.

\bibitem[Ma et~al.(2022)Ma, Farshidian, Miki, Lee, and Hutter]{ma2022combining}
Yuntao Ma, Farbod Farshidian, Takahiro Miki, Joonho Lee, and Marco Hutter.
\newblock Combining learning-based locomotion policy with model-based manipulation for legged mobile manipulators.
\newblock \emph{IEEE Robotics and Automation Letters}, 7\penalty0 (2):\penalty0 2377--2384, 2022.
\newblock \doi{10.1109/LRA.2022.3143567}.

\bibitem[Mandlekar et~al.(2018)Mandlekar, Zhu, Garg, Booher, Spero, Tung, Gao, Emmons, Gupta, Orbay, et~al.]{mandlekar2018roboturk}
Ajay Mandlekar, Yuke Zhu, Animesh Garg, Jonathan Booher, Max Spero, Albert Tung, Julian Gao, John Emmons, Anchit Gupta, Emre Orbay, et~al.
\newblock Roboturk: A crowdsourcing platform for robotic skill learning through imitation.
\newblock In \emph{Conference on Robot Learning}, pages 879--893. PMLR, 2018.

\bibitem[Meeker et~al.(2020)Meeker, Haas-Heger, and Ciocarlie]{meeker2020continuous}
Cassie Meeker, Maximilian Haas-Heger, and Matei Ciocarlie.
\newblock A continuous teleoperation subspace with empirical and algorithmic mapping algorithms for nonanthropomorphic hands.
\newblock \emph{IEEE Transactions on Automation Science and Engineering}, 19\penalty0 (1):\penalty0 373--386, 2020.

\bibitem[Mosbach et~al.(2022)Mosbach, Moraw, and Behnke]{mosbach2022accelerating}
Malte Mosbach, Kara Moraw, and Sven Behnke.
\newblock Accelerating interactive human-like manipulation learning with gpu-based simulation and high-quality demonstrations.
\newblock In \emph{2022 IEEE-RAS 21st International Conference on Humanoid Robots (Humanoids)}, pages 435--441. IEEE, 2022.

\bibitem[Nair et~al.(2020)Nair, Gupta, Dalal, and Levine]{nair2020awac}
Ashvin Nair, Abhishek Gupta, Murtaza Dalal, and Sergey Levine.
\newblock Awac: Accelerating online reinforcement learning with offline datasets.
\newblock \emph{arXiv preprint arXiv:2006.09359}, 2020.

\bibitem[Ng et~al.(2000)Ng, Russell, et~al.]{Ng2000}
Andrew~Y Ng, Stuart~J Russell, et~al.
\newblock Algorithms for inverse reinforcement learning.
\newblock In \emph{ICML}, 2000.

\bibitem[{Octo Model Team} et~al.(2023){Octo Model Team}, Ghosh, Walke, Pertsch, Black, Mees, Dasari, Hejna, Xu, Luo, Kreiman, Tan, Sadigh, Finn, and Levine]{octo_2023}
{Octo Model Team}, Dibya Ghosh, Homer Walke, Karl Pertsch, Kevin Black, Oier Mees, Sudeep Dasari, Joey Hejna, Charles Xu, Jianlan Luo, Tobias Kreiman, {You Liang} Tan, Dorsa Sadigh, Chelsea Finn, and Sergey Levine.
\newblock Octo: An open-source generalist robot policy.
\newblock \url{https://octo-models.github.io}, 2023.

\bibitem[Padalkar et~al.(2023)Padalkar, Pooley, Jain, Bewley, Herzog, Irpan, Khazatsky, Rai, Singh, Brohan, et~al.]{padalkar2023open}
Abhishek Padalkar, Acorn Pooley, Ajinkya Jain, Alex Bewley, Alex Herzog, Alex Irpan, Alexander Khazatsky, Anant Rai, Anikait Singh, Anthony Brohan, et~al.
\newblock Open x-embodiment: Robotic learning datasets and rt-x models.
\newblock \emph{arXiv preprint arXiv:2310.08864}, 2023.

\bibitem[Pari et~al.(2021)Pari, Shafiullah, Arunachalam, and Pinto]{pari2021surprising}
Jyothish Pari, Nur~Muhammad Shafiullah, Sridhar~Pandian Arunachalam, and Lerrel Pinto.
\newblock The surprising effectiveness of representation learning for visual imitation, 2021.

\bibitem[Pomerleau(1988)]{ALVINN}
Dean~A. Pomerleau.
\newblock Alvinn: An autonomous land vehicle in a neural network.
\newblock In D.~Touretzky, editor, \emph{NeurIPS}, volume~1. Morgan-Kaufmann, 1988.

\bibitem[Qin et~al.(2022)Qin, Su, and Wang]{qin2022one}
Yuzhe Qin, Hao Su, and Xiaolong Wang.
\newblock From one hand to multiple hands: Imitation learning for dexterous manipulation from single-camera teleoperation.
\newblock \emph{arXiv preprint arXiv:2204.12490}, 2022.

\bibitem[Qin et~al.(2023)Qin, Yang, Huang, Van~Wyk, Su, Wang, Chao, and Fox]{qin2023anyteleop}
Yuzhe Qin, Wei Yang, Binghao Huang, Karl Van~Wyk, Hao Su, Xiaolong Wang, Yu-Wei Chao, and Dietor Fox.
\newblock Anyteleop: A general vision-based dexterous robot arm-hand teleoperation system.
\newblock \emph{arXiv preprint arXiv:2307.04577}, 2023.

\bibitem[Radosavovic et~al.(2022)Radosavovic, Xiao, James, Abbeel, Malik, and Darrell]{MVP}
Ilija Radosavovic, Tete Xiao, Stephen James, Pieter Abbeel, Jitendra Malik, and Trevor Darrell.
\newblock Real-world robot learning with masked visual pre-training, 2022.
\newblock URL \url{https://arxiv.org/abs/2210.03109}.

\bibitem[Reuss et~al.(2023)Reuss, Li, Jia, and Lioutikov]{reuss2023goal}
Moritz Reuss, Maximilian Li, Xiaogang Jia, and Rudolf Lioutikov.
\newblock Goal-conditioned imitation learning using score-based diffusion policies.
\newblock \emph{arXiv preprint arXiv:2304.02532}, 2023.

\bibitem[Reynolds et~al.(2009)]{reynolds2009gaussian}
Douglas~A Reynolds et~al.
\newblock Gaussian mixture models.
\newblock \emph{Encyclopedia of biometrics}, 741\penalty0 (659-663), 2009.

\bibitem[Schwarz et~al.()Schwarz, Lenz, Rochow, Schreiber, and Behnke]{schwarznimbro}
Max Schwarz, Christian Lenz, Andre Rochow, Michael Schreiber, and Sven Behnke.
\newblock Nimbro avatar: Interactive immersive telepresence with force-feedback telemanipulation. in 2021 ieee.
\newblock In \emph{RSJ International Conference on Intelligent Robots and Systems (IROS)}, pages 5312--5319.

\bibitem[Shafiullah et~al.(2022)Shafiullah, Cui, Altanzaya, and Pinto]{shafiullah2022behavior}
Nur~Muhammad Shafiullah, Zichen Cui, Ariuntuya~Arty Altanzaya, and Lerrel Pinto.
\newblock Behavior transformers: Cloning $ k $ modes with one stone.
\newblock \emph{Advances in neural information processing systems}, 35:\penalty0 22955--22968, 2022.

\bibitem[Shafiullah et~al.(2023)Shafiullah, Rai, Etukuru, Liu, Misra, Chintala, and Pinto]{dobbe}
Nur Muhammad~Mahi Shafiullah, Anant Rai, Haritheja Etukuru, Yiqian Liu, Ishan Misra, Soumith Chintala, and Lerrel Pinto.
\newblock On bringing robots home.
\newblock \emph{arXiv preprint arXiv:2311.16098}, 2023.

\bibitem[Shah et~al.(2023)Shah, Sridhar, Dashora, Stachowicz, Black, Hirose, and Levine]{shah2023vint}
Dhruv Shah, Ajay Sridhar, Nitish Dashora, Kyle Stachowicz, Kevin Black, Noriaki Hirose, and Sergey Levine.
\newblock Vi{NT}: A foundation model for visual navigation.
\newblock In \emph{7th Annual Conference on Robot Learning}, 2023.
\newblock URL \url{https://arxiv.org/abs/2306.14846}.

\bibitem[Sian et~al.(2004)Sian, Yokoi, Kajita, Kanehiro, and Tanie]{sian2004whole}
Neo~Ee Sian, Kazuhito Yokoi, Shuuji Kajita, Fumio Kanehiro, and Kazuo Tanie.
\newblock Whole body teleoperation of a humanoid robot development of a simple master device using joysticks.
\newblock \emph{Journal of the Robotics Society of Japan}, 22\penalty0 (4):\penalty0 519--527, 2004.

\bibitem[Sivakumar et~al.(2022)Sivakumar, Shaw, and Pathak]{sivakumar2022robotic}
Aravind Sivakumar, Kenneth Shaw, and Deepak Pathak.
\newblock Robotic telekinesis: Learning a robotic hand imitator by watching humans on youtube, 2022.

\bibitem[Smith et~al.(2022)Smith, Kostrikov, and Levine]{smith2022walk}
Laura Smith, Ilya Kostrikov, and Sergey Levine.
\newblock A walk in the park: Learning to walk in 20 minutes with model-free reinforcement learning.
\newblock \emph{arXiv preprint arXiv:2208.07860}, 2022.

\bibitem[Song et~al.(2020)Song, Zeng, Lee, and Funkhouser]{song2020grasping}
Shuran Song, Andy Zeng, Johnny Lee, and Thomas Funkhouser.
\newblock Grasping in the wild: Learning 6dof closed-loop grasping from low-cost demonstrations.
\newblock \emph{RA-L}, 2020.

\bibitem[{Stanford Artificial Intelligence Laboratory et al.}()]{ros}
{Stanford Artificial Intelligence Laboratory et al.}
\newblock Robotic operating system.
\newblock URL \url{https://www.ros.org}.

\bibitem[Wang et~al.(2023)Wang, Fan, Sun, Zhang, Fei-Fei, Xu, Zhu, and Anandkumar]{wang2023mimicplay}
Chen Wang, Linxi Fan, Jiankai Sun, Ruohan Zhang, Li~Fei-Fei, Danfei Xu, Yuke Zhu, and Anima Anandkumar.
\newblock Mimicplay: Long-horizon imitation learning by watching human play.
\newblock \emph{arXiv preprint arXiv:2302.12422}, 2023.

\bibitem[Wu et~al.(2023)Wu, Shentu, Yi, Lin, and Abbeel]{wu2023gello}
Philipp Wu, Yide Shentu, Zhongke Yi, Xingyu Lin, and Pieter Abbeel.
\newblock Gello: A general, low-cost, and intuitive teleoperation framework for robot manipulators.
\newblock \emph{arXiv preprint arXiv:2309.13037}, 2023.

\bibitem[Wu et~al.(2019)Wu, Balatti, Lorenzini, Zhao, Kim, and Ajoudani]{wu2019teleoperation}
Yuqiang Wu, Pietro Balatti, Marta Lorenzini, Fei Zhao, Wansoo Kim, and Arash Ajoudani.
\newblock A teleoperation interface for loco-manipulation control of mobile collaborative robotic assistant.
\newblock \emph{IEEE Robotics and Automation Letters}, 4\penalty0 (4):\penalty0 3593--3600, 2019.

\bibitem[xArm Developer()]{xArmPythonSDK}
xArm Developer.
\newblock xarm python sdk.
\newblock \url{https://github.com/xArm-Developer/xArm-Python-SDK}.

\bibitem[Zhang et~al.(2020)Zhang, Bazarevsky, Vakunov, Tkachenka, Sung, Chang, and Grundmann]{zhang2020mediapipe}
Fan Zhang, Valentin Bazarevsky, Andrey Vakunov, Andrei Tkachenka, George Sung, Chuo-Ling Chang, and Matthias Grundmann.
\newblock Mediapipe hands: On-device real-time hand tracking, 2020.

\bibitem[Zhang et~al.(2016)Zhang, Kahn, Levine, and Abbeel]{zhang2016learning}
Tianhao Zhang, Gregory Kahn, Sergey Levine, and Pieter Abbeel.
\newblock Learning deep control policies for autonomous aerial vehicles with mpc-guided policy search.
\newblock In \emph{2016 IEEE international conference on robotics and automation (ICRA)}, pages 528--535. IEEE, 2016.

\bibitem[Zhang et~al.(2018)Zhang, McCarthy, Jow, Lee, Chen, Goldberg, and Abbeel]{zhang2018deep}
Tianhao Zhang, Zoe McCarthy, Owen Jow, Dennis Lee, Xi~Chen, Ken Goldberg, and Pieter Abbeel.
\newblock Deep imitation learning for complex manipulation tasks from virtual reality teleoperation.
\newblock In \emph{ICRA}, 2018.

\bibitem[Zhao et~al.(2023)Zhao, Kumar, Levine, and Finn]{zhao2023learning}
Tony~Z Zhao, Vikash Kumar, Sergey Levine, and Chelsea Finn.
\newblock Learning fine-grained bimanual manipulation with low-cost hardware.
\newblock \emph{arXiv preprint arXiv:2304.13705}, 2023.

\bibitem[Zhao et~al.(2012)Zhao, Chai, and Xu]{zhao2012combining}
Wenping Zhao, Jinxiang Chai, and Ying-Qing Xu.
\newblock Combining marker-based mocap and rgb-d camera for acquiring high-fidelity hand motion data.
\newblock In \emph{Proceedings of the ACM SIGGRAPH/eurographics symposium on computer animation}, pages 33--42, 2012.

\bibitem[Zhu et~al.(2022)Zhu, Joshi, Stone, and Zhu]{zhu2022viola}
Yifeng Zhu, Abhishek Joshi, Peter Stone, and Yuke Zhu.
\newblock Viola: Imitation learning for vision-based manipulation with object proposal priors.
\newblock \emph{arXiv preprint arXiv:2210.11339}, 2022.
\newblock \doi{10.48550/arXiv.2210.11339}.

\end{thebibliography}

\clearpage
\appendix

\label{appendix}

\subsection{Framework details}
\label{appendix:subsec:framework}

\subsubsection{Structure of the framework}
We use ZeroMQ for networking between nodes. The \method{} framework is divided into two parts - \textit{teleoperation} and \textit{data collection}.

\mysection{Teleoperation} The teleoperator is divided into 5 components - Detector, Keypoint Transformer, Operator, Controller, and Visualizer. A brief description of each  has been provided below.

\begin{enumerate}
    \item \textbf{Detector:} Receives the hand keypoints from the Meta Quest 3 and publishes them to ZMQ sockets.
    \item \textbf{Keypoint Transformer:} Subscribes the keypoints published by the detector and maps them to the robot pose.
    \item \textbf{Operator:} Receives the robot pose from the keypoint transformer and the current robot state from the controller. The operator computes the robot's actions which are published to a ZMQ socket.
    \item \textbf{Controller:} Subscribes an action from the operator and takes an action in the real or simulated environment. After taking the action, the controller publishes the current states of the environment for use by the operator.
    \item \textbf{Visualizer:} Subscribes the RGB images from the camera process (or the environment in case of simulations) and puts it on the screen inside the VR app for visualization during teleoperation.
\end{enumerate}

\mysection{Data Collection} A data recorder process subscribes sensor information (RGB and Depth images, tactile readings, timestamps) and robot-specific information (joint states, gripper states, timestamps) from the corresponding sockets and logs them in corresponding files. The data is then compiled together by matching the timestamps between the sensor information and robot-specific data.

\begin{figure*}[!t]
    \centering
    \includegraphics[width=0.6\linewidth]{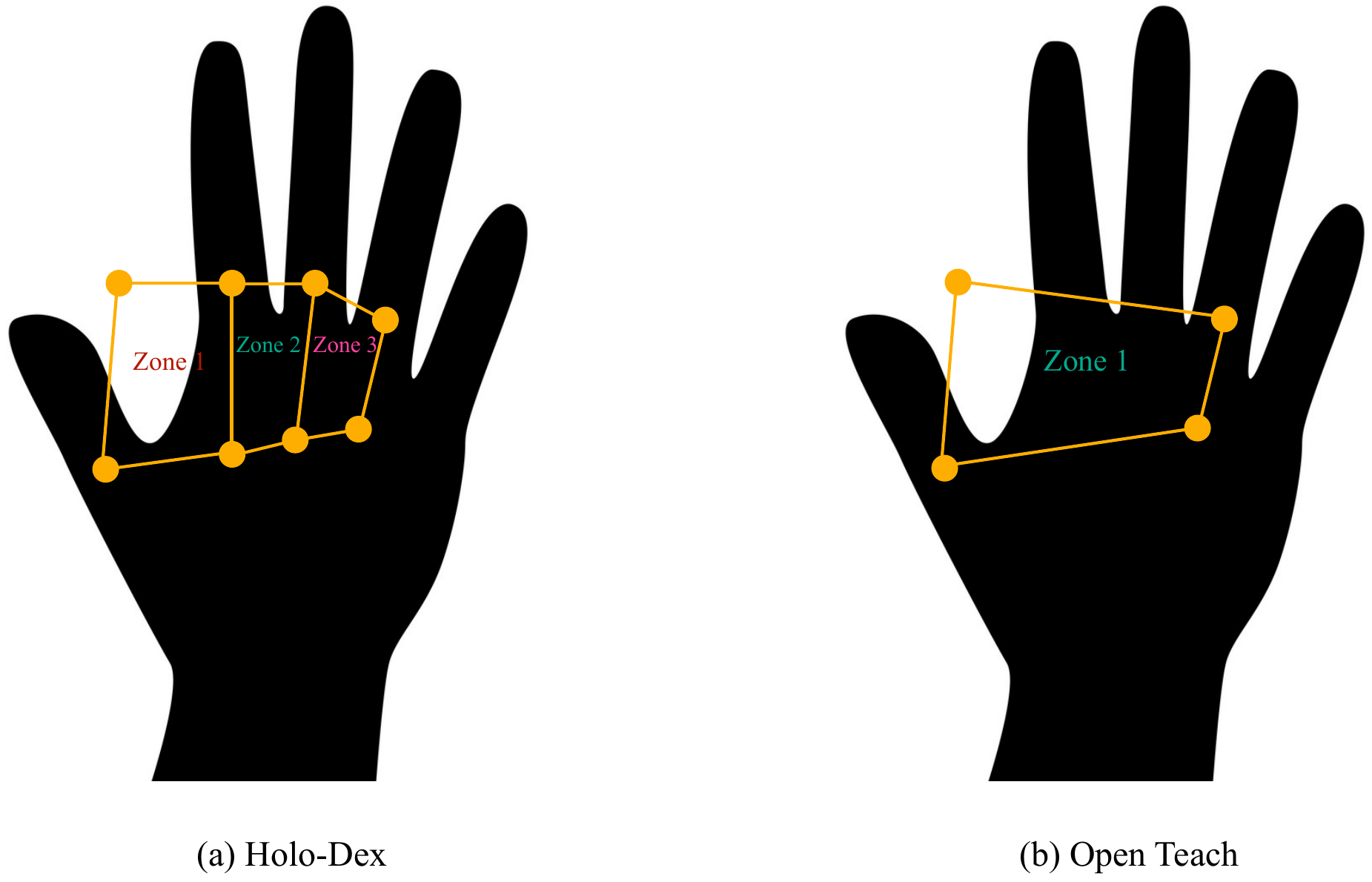}
    \caption{Thumb retargeting difference}
    \label{appendix:fig:thumb_retargeting}
\end{figure*}

\subsubsection{Thumb Retargeting for Robot Hand}
\label{appendix:subsec:thumb_retargeting}
Section~\ref{subsec:retargeting} provides details about the design of the \method{} wrapper for the robot hand. To recap, given the individual joint angles in the teacher's hand from the VR headset, the joint angles for the robot hand can be computed by directly commanding the robot's joints to the corresponding angles. This works well for all fingers except the thumb. Holo-Dex\cite{arunachalam2023holo} deals with this by mapping the spatial coordinate of the teacher’s thumb tip to that of the robot hand.
Then an inverse kinematics solver is used to compute the joint angles of the thumb. In this case, the retargeting of the thumb is done in 2D space. These bounds, depicted in Fig.~\ref{appendix:fig:thumb_retargeting}(a), define the thumb's reach limits. During retargeting, the thumb tip's zone on the 2D palm plane is detected, and a perspective transform from the human hand to the robot hand is applied, aligning the human thumb tip with the robot thumb tip on the 2D plane. However, using three separate bounds introduces jitters when the thumb tip transitions between zones and results in stagnancy when outside the bounds. Further, in Holo-Dex, the height of the robot thumb tip is fixed, allowing it to only move along the 2D space.

To address these challenges, \method{} employs a single, large zone spanning the entire thumb's workspace in 2D space(refer to Fig.~\ref{appendix:fig:thumb_retargeting}(b)). When the thumb is within bounds, a perspective transformation retargets the human thumb tip to the robot thumb tip. In cases where the thumb goes out of bounds, the closest point within the bound is estimated and used for retargeting, avoiding stagnation. Additionally, instead of a fixed height, the thumb is allowed to move perpendicular to the 2D surface along the palm, mapping the height of the human thumb tip to the robot thumb tip based on maximum and minimum height bounds. This approach ensures smoother thumb motion and enables the performance of more complex tasks compared to Holo-Dex~\cite{arunachalam2023holo}.



\begin{table}[!th]
\centering
\caption{Time }
\label{appendix:table:task_times}
\begin{tabular}{llc}
\hline
\rowcolor[HTML]{FFFFFF} 
\textbf{Robot Setup} & \textbf{Task}                                                                     & \textbf{\begin{tabular}[c]{@{}c@{}}Average time to\\ collect a demo\\ (in s)\end{tabular}} \\ \hline
\rowcolor[HTML]{EFEFEF} 
Franka-Allegro       & Open box                                                                          & 45                                                                                       \\
\rowcolor[HTML]{EFEFEF} 
                     & Grasp sponge                                                                      & 60                                                                                         \\
\rowcolor[HTML]{EFEFEF} 
                     & Pick up tea satchet                                                               & 60                                                                                       \\
\rowcolor[HTML]{EFEFEF} 
                     & Grasp object and twist                                                            & 35                                                                                       \\
\rowcolor[HTML]{FFFFFF} 
Kinova-Allegro       & Unfold towel                                                                      & 40                                                                                       \\
\rowcolor[HTML]{FFFFFF} 
                     & Open a pack of cream                                                              & 10                                                                                       \\
\rowcolor[HTML]{FFFFFF} 
                     & Open ketchup bottle                                                               & 40                                                                                       \\
\rowcolor[HTML]{EFEFEF} 
Bimanual             & Uncap marker                                                                      & 60                                                                                       \\
\rowcolor[HTML]{EFEFEF} 
                     & Sweep table                                                                       & 60                                                                                       \\
\rowcolor[HTML]{EFEFEF} 
                     & Pour sprinkles in a bowl                                                          & 40                                                                                       \\
\rowcolor[HTML]{FFFFFF} 
Allegro Sim          & Flip cube                                                                         & 3                                                                                        \\
\rowcolor[HTML]{FFFFFF} 
                     & Flip sponge                                                                       & 20                                                                                       \\
\rowcolor[HTML]{FFFFFF} 
                     & Pinch Grasp                                                                       & 15                                                                                       \\
\rowcolor[HTML]{EFEFEF} 
LIBERO Sim           & Close top drawer of cabinet                                                       & 10                                                                                       \\
\rowcolor[HTML]{EFEFEF} 
                     & Turn on stove                                                                     & 25                                                                                       \\
\rowcolor[HTML]{EFEFEF} 
                     & \begin{tabular}[l]{@{}l@{}}Pick up and put soup \\ can in the basket\end{tabular} & 30                                                                                       \\ \hline
\end{tabular}
\end{table}

\subsection{Task Details}
\label{appendix:subsec:robot_task_details}

\subsubsection{Demo Collections times}
\label{appendix:subsubsec:demo_collect_times}
Table~\ref{appendix:table:task_times} provides the average times required to collect a demonstration for 16 tasks across 3 real-world setups (Franka-Allegro, Kinova-Allegro, Bimanual) and 2 simulated environments(Allegro sim, LIBERO sim).

\subsubsection{Task Descriptions}
\label{appendix:subsubsec:task_desc}
\begin{figure*}[!th]
    \centering
    \includegraphics[width=0.95\textwidth]{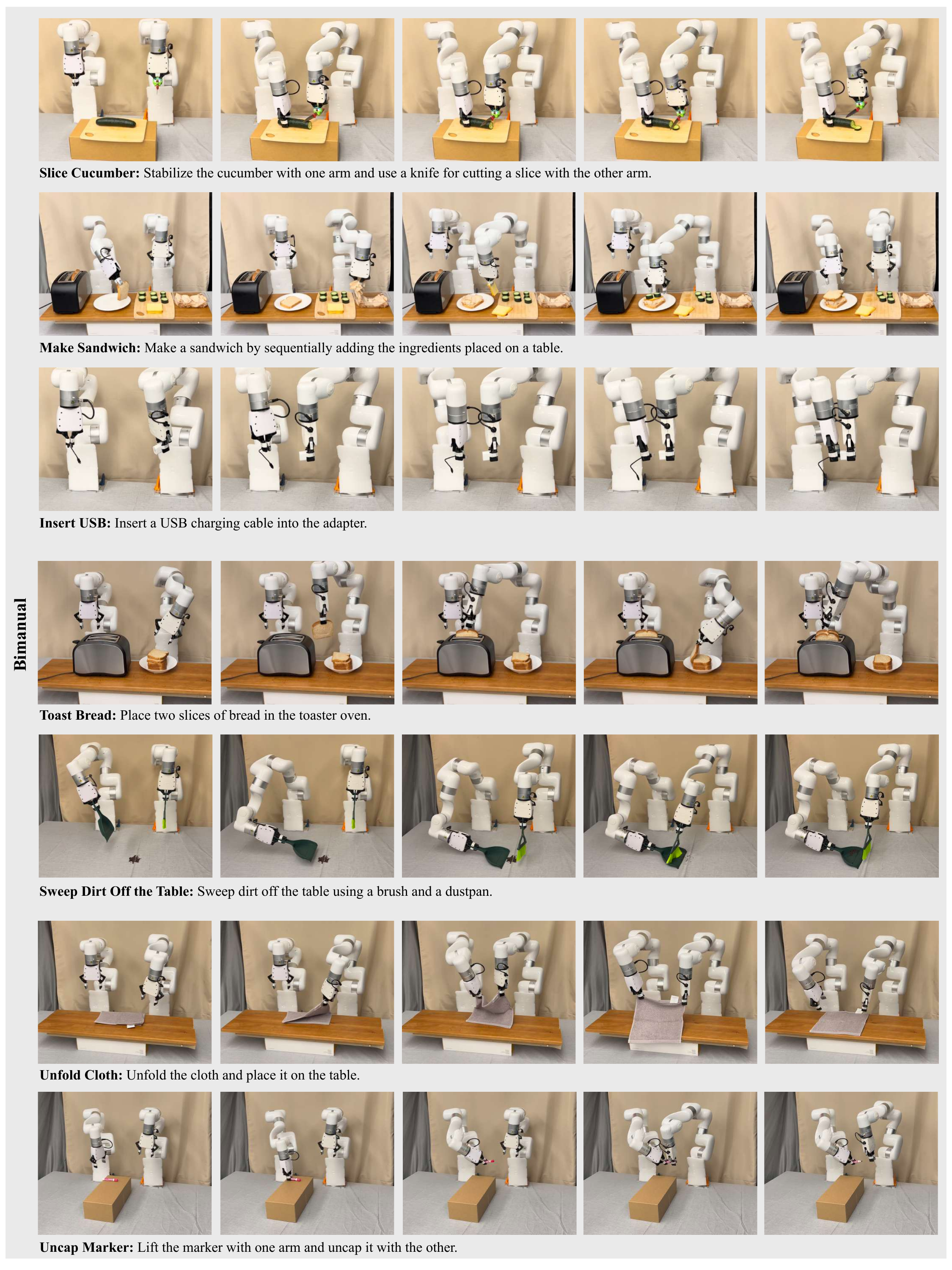}
    \caption{Real world task rollouts demonstrating the ability of \method{} to perform intricate, long-horizon tasks.}
    \label{appendix:fig:task_rollouts1}
\end{figure*}

\begin{figure*}[!th]
    \centering
    \includegraphics[width=0.95\textwidth]{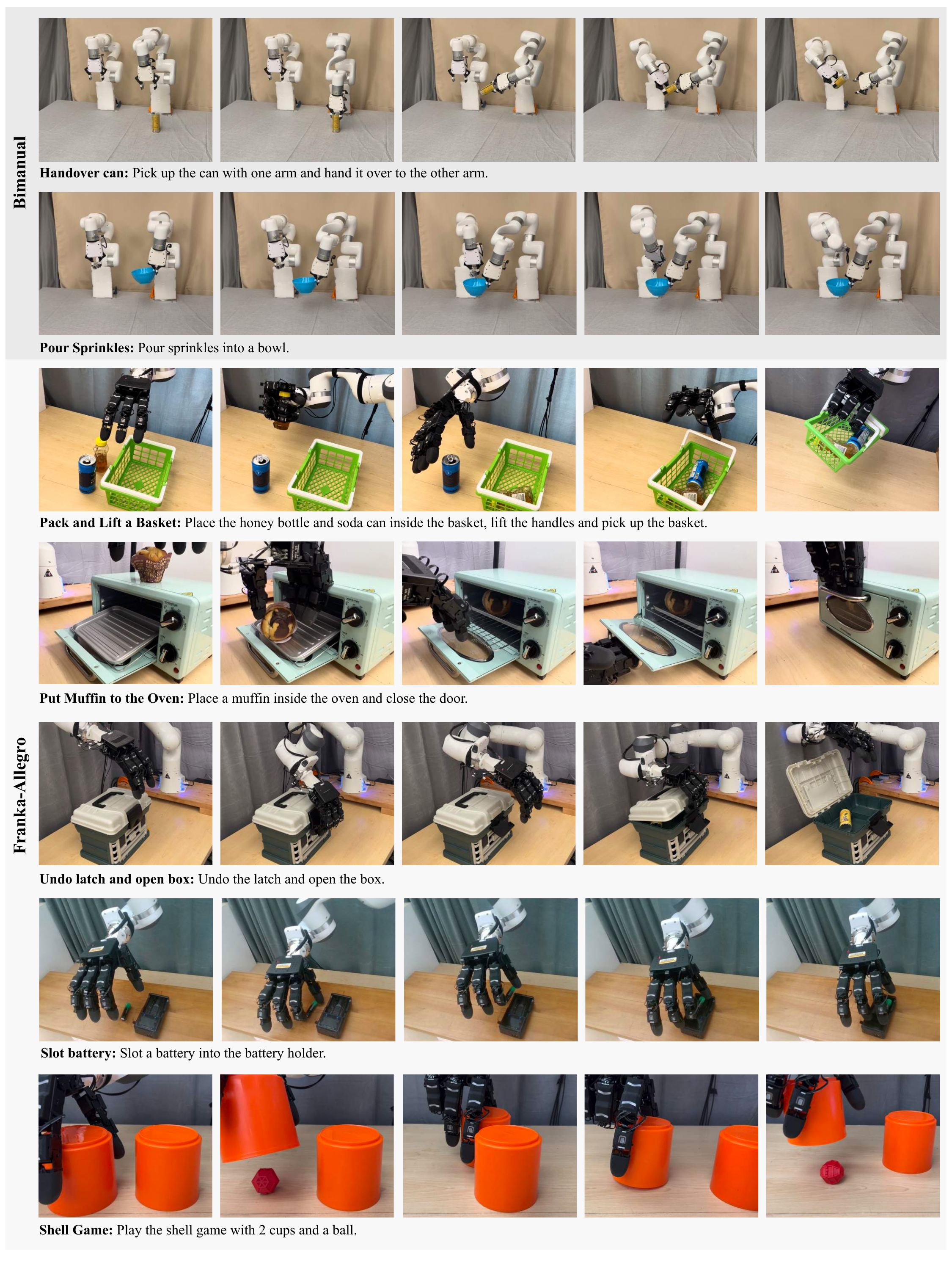}
    \caption{Real world task rollouts demonstrating the ability of \method{} to perform intricate, long-horizon tasks.}
    \label{appendix:fig:task_rollouts2}
\end{figure*}

\begin{figure*}[!th]
    \centering
    \includegraphics[width=0.95\textwidth]{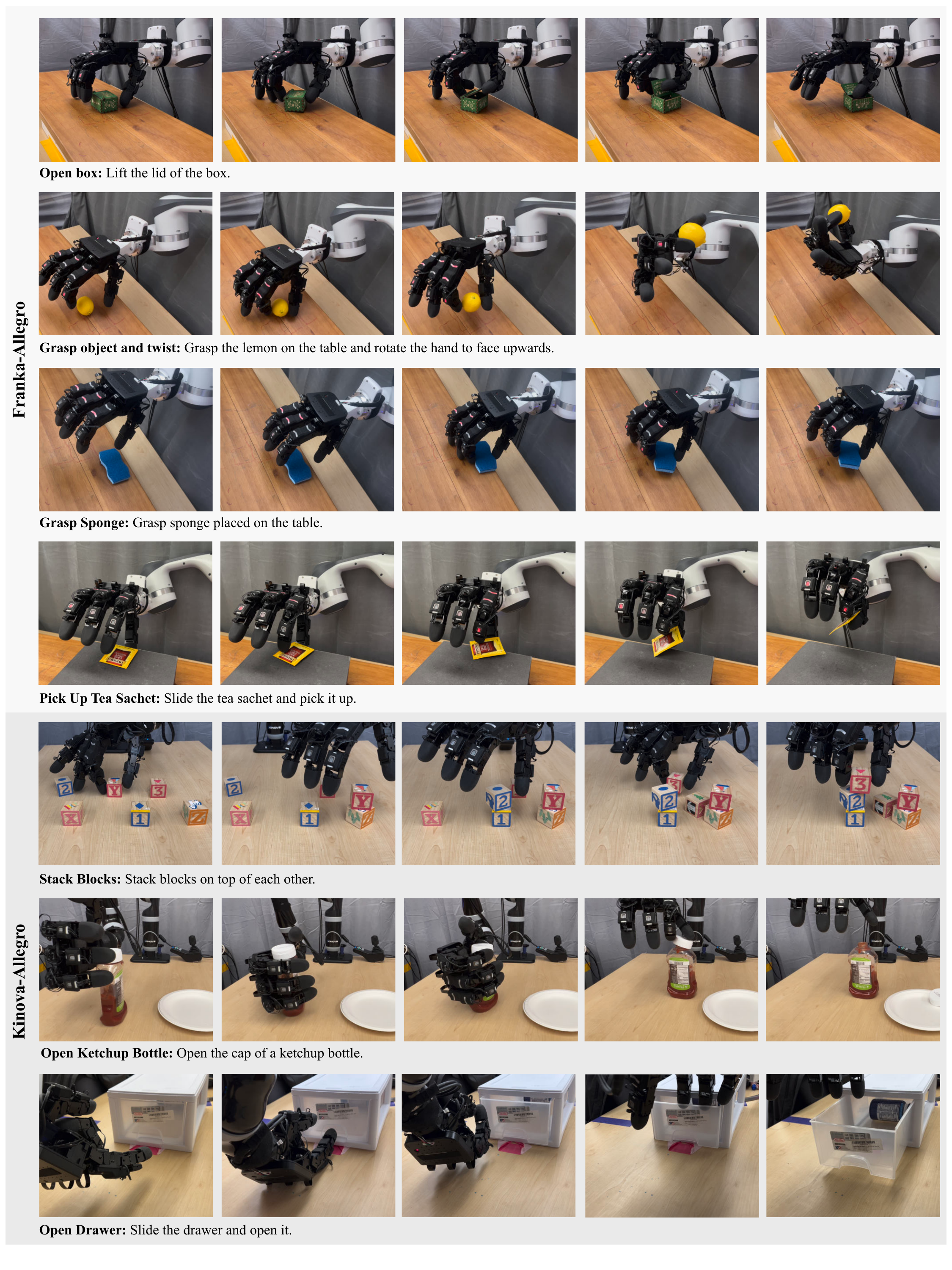}
    \caption{Real world task rollouts demonstrating the ability of \method{} to perform intricate, long-horizon tasks.}
    \label{appendix:fig:task_rollouts3}
\end{figure*}

\begin{figure*}[!th]
    \centering
    \includegraphics[width=0.95\textwidth]{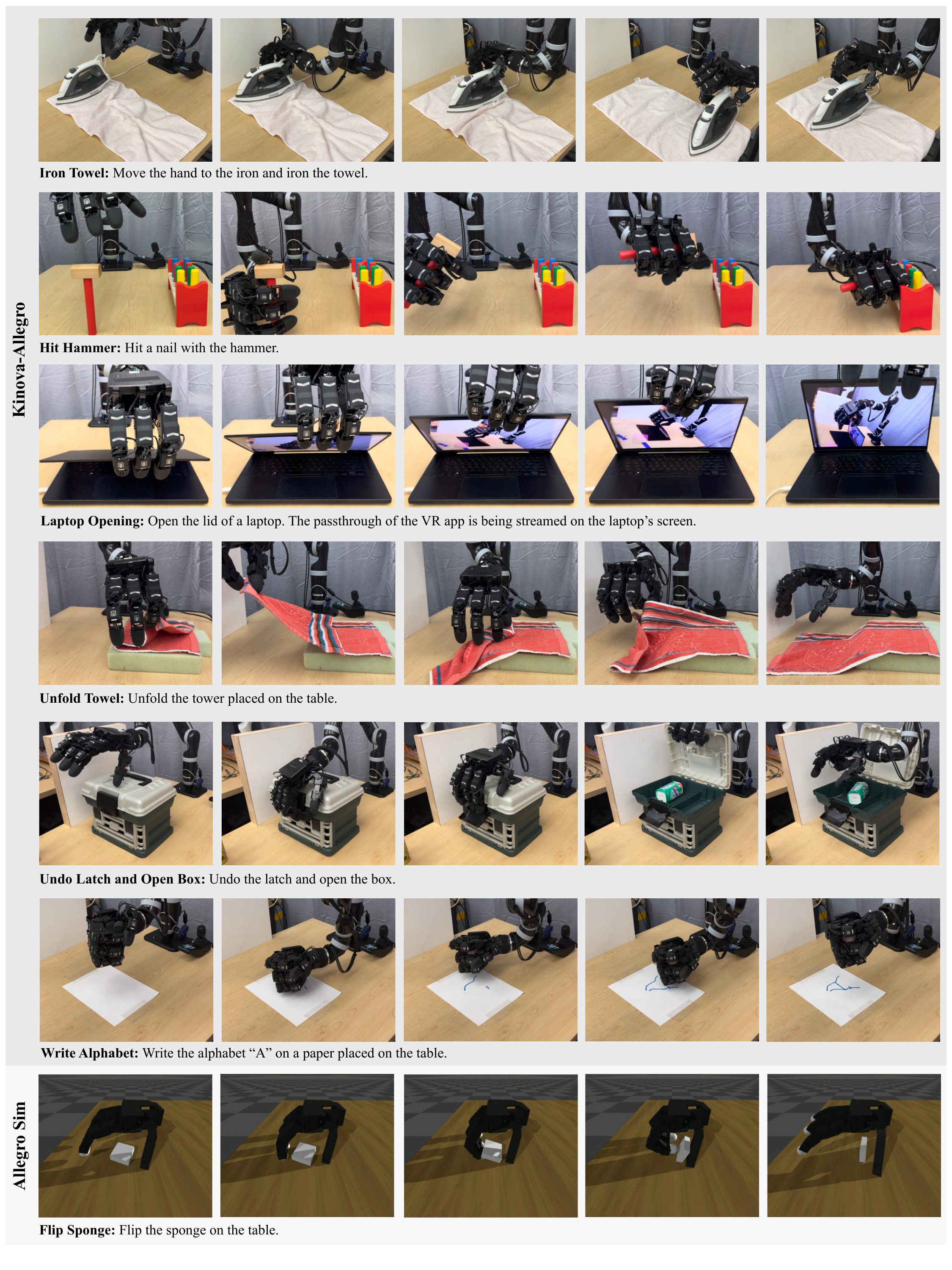}
    \caption{Real world task rollouts demonstrating the ability of \method{} to perform intricate, long-horizon tasks.}
    \label{appendix:fig:task_rollouts4}
\end{figure*}

\begin{figure*}[!th]
    \centering
    \includegraphics[width=0.95\textwidth]{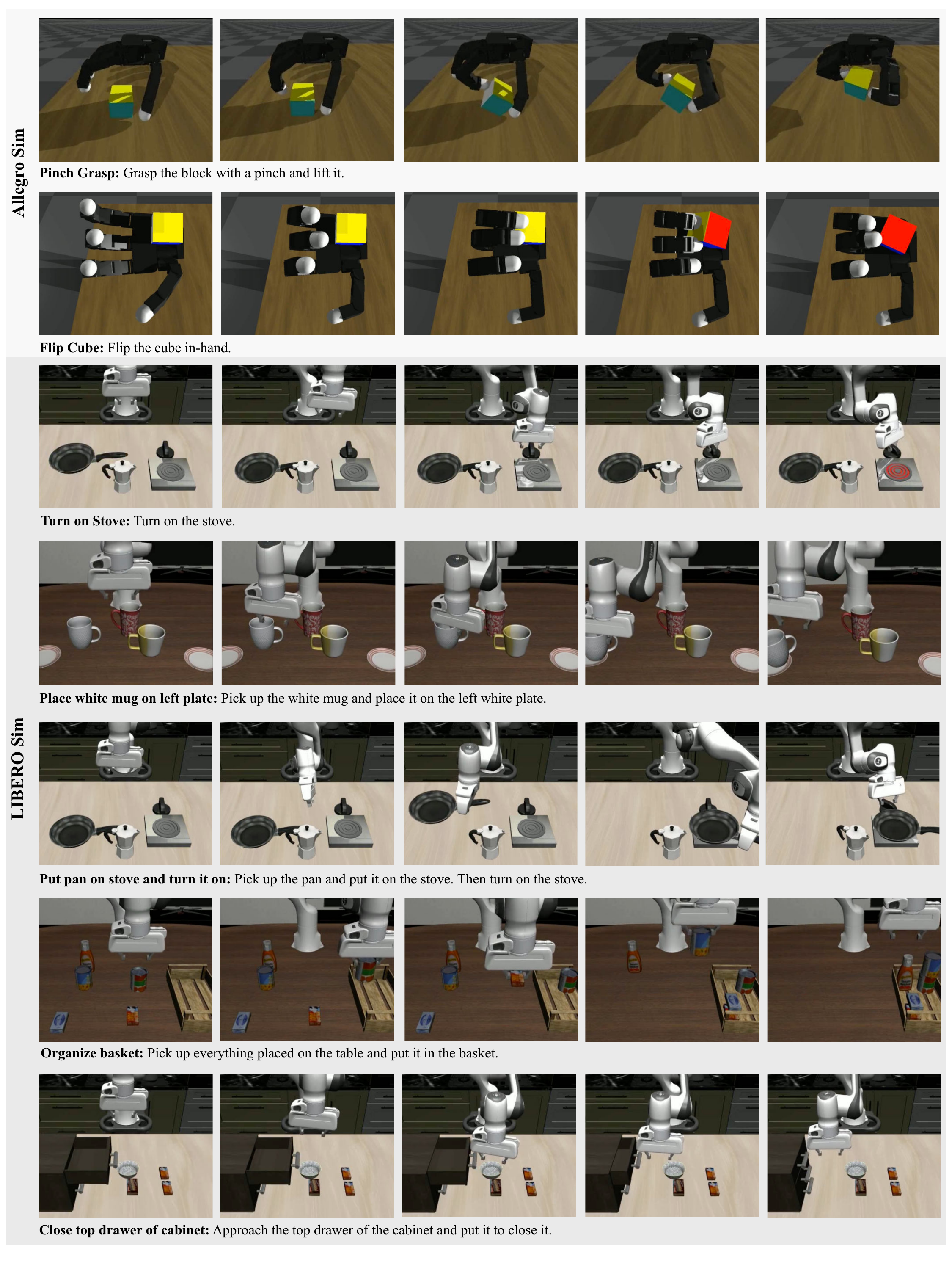}
    \caption{Real world task rollouts demonstrating the ability of \method{} to perform intricate, long-horizon tasks.}
    \label{appendix:fig:task_rollouts5}
\end{figure*}

\begin{figure*}[!th]
    \centering
    \includegraphics[width=0.95\textwidth]{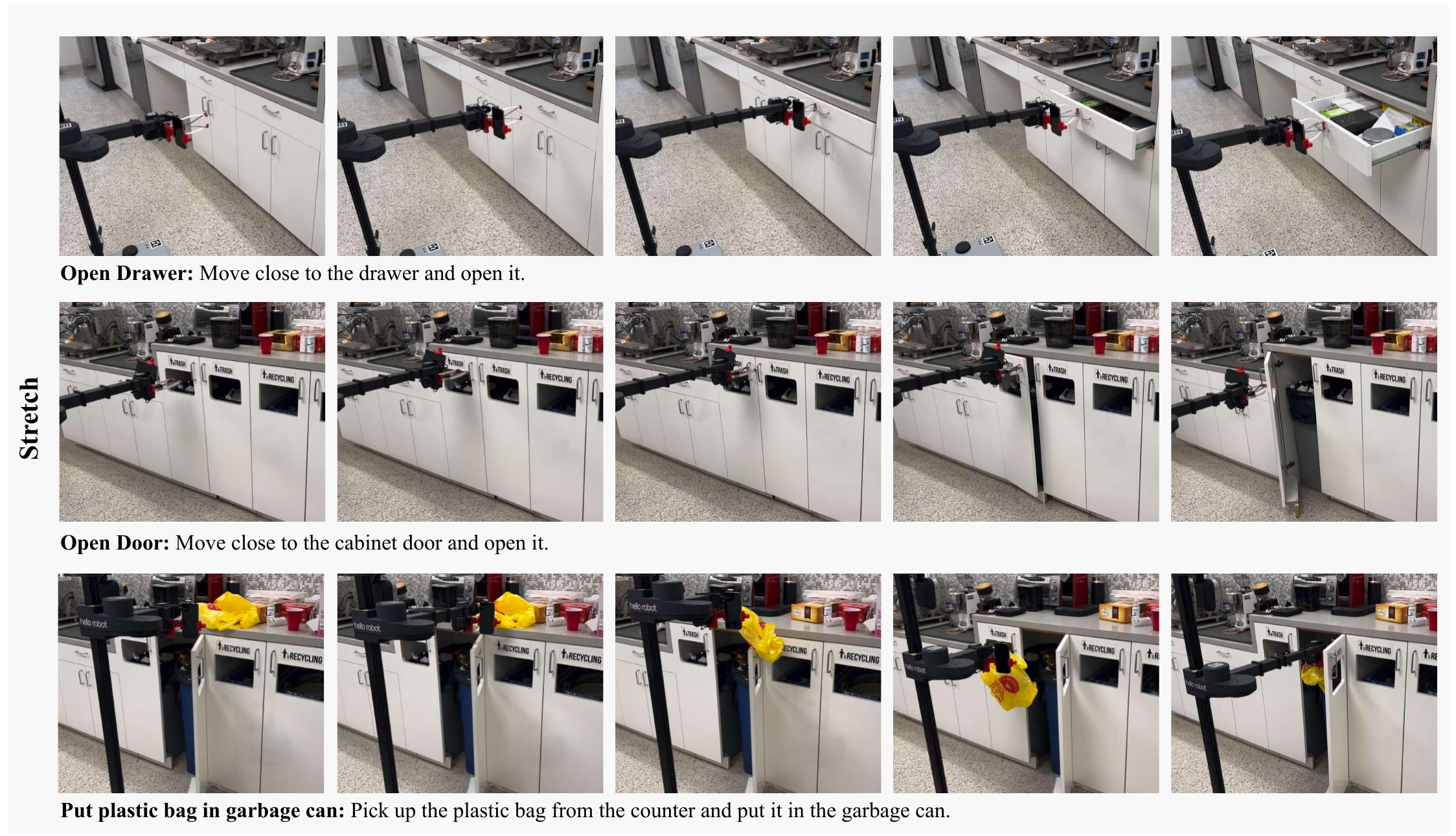}
    \caption{Real world task rollouts demonstrating the ability of \method{} to perform intricate, long-horizon tasks.}
    \label{appendix:fig:task_rollouts6}
\end{figure*}

Fig.~\ref{appendix:fig:task_rollouts1}, Fig.~\ref{appendix:fig:task_rollouts2}, Fig.~\ref{appendix:fig:task_rollouts3}, Fig.~\ref{appendix:fig:task_rollouts4}, Fig.~\ref{appendix:fig:task_rollouts5}, and Fig.~\ref{appendix:fig:task_rollouts6} provide rollouts of all the tasks performed both in the real world and in simulated environments. Each task rollout is labeled with the name of the task and a task description.

\subsection{User Study}
\label{appendix:user_study}

Following up from Section~\ref{subsec:user_study}, we provide the success rate and average completion times for all 15 users for each task performed in Table~\ref{appendix:table:user_study_succ} and Table~\ref{appendix:table:user_study_time} respectively. Each user roughly performed 3 tasks on average, with 5 trials for each task. As mentioned in Section~\ref{subsec:user_study}, since the Holo-Dex~\cite{arunachalam2023holo} and AnyTeleop~\cite{qin2023anyteleop} baselines lack open-source code for arm retargeting, we were unable to evaluate them on tasks involving arm movements. We observe a wide range of differences in success rates and average completion times demonstrating the inherent variations across users.

\begin{table*}[!th]
\centering
\caption{Success rates for the user study conducted across 15 individuals. Each user roughly performs 3 tasks on average.}
\label{appendix:table:user_study_succ}
\begin{tabular}{ccccccc}
\hline
\textbf{User} & \textbf{Method} & \multicolumn{5}{c}{\textbf{Success Rate (in 5 trials)}}                                                                        \\ \cline{3-7}
\textbf{}     & \textbf{}       & \textbf{Flip Cube} & \textbf{Pinch Grasp} & \textbf{Pour} & \textbf{Pick and Place} & \textbf{Open Box of Mints} \\ \hline
\rowcolor[HTML]{EFEFEF} 
User 1        & Holo-Dex        & 1                  & 0                    & -             & -                       & -                          \\
\rowcolor[HTML]{EFEFEF} 
              & AnyTeleop       & 0.8                & 0.2                  & -             & -                       & -                          \\
\rowcolor[HTML]{EFEFEF} 
              & Open Teach      & 1                  & 0.8                  & 0.2           & -                       & -                          \\
User 2        & Holo-Dex        & -                  & 0.2                  & -             & -                       & -                          \\
              & AnyTeleop       & -                  & 0.2                  & -             & -                       & -                          \\
              & Open Teach      & -                  & 0.8                  & -             & 0.8                     & 0.8                        \\
\rowcolor[HTML]{EFEFEF} 
User 3        & Holo-Dex        & 1                  & 0                    & -             & -                       & -                          \\
\rowcolor[HTML]{EFEFEF} 
              & AnyTeleop       & 1                  & 0.2                  & -             & -                       & -                          \\
\rowcolor[HTML]{EFEFEF} 
              & Open Teach      & 1                  & 0.8                  & -             & -                       & 0.2                        \\
User 4       & Holo-Dex        & 1                  & 0                    & -             & -                       & -                          \\
              & AnyTeleop       & 1                  & 0.2                  & -             & -                       & -                          \\
              & Open Teach      & 1                  & 0.8                  & -             & 0.6                     & 0.4                        \\
\rowcolor[HTML]{EFEFEF} 
User 5       & Holo-Dex        & -                  & 0                    & -             & -                       & -                          \\
\rowcolor[HTML]{EFEFEF} 
              & AnyTeleop       & -                  & 0.6                  & -             & -                       & -                          \\
\rowcolor[HTML]{EFEFEF} 
              & Open Teach      & -                  & 0.2                  & 0.4           & 1                       & -                          \\
User 6        & Holo-Dex        & -                  & 0                    & -             & -                       & -                          \\
              & AnyTeleop       & -                  & 0.6                  & -             & -                       & -                          \\
              & Open Teach      & -                  & 0.8                  & -             & 0.2                     & -                          \\
\rowcolor[HTML]{EFEFEF} 
User 7       & Holo-Dex        & -                  & 0                    & -             & -                       & -                          \\
\rowcolor[HTML]{EFEFEF} 
              & AnyTeleop       & -                  & 0                    & -             & -                       & -                          \\
\rowcolor[HTML]{EFEFEF} 
              & Open Teach      & -                  & 0.6                  & 0.8           & 0.8                     & 0.4                        \\
User 8       & Holo-Dex        & 1                  & -                    & -             & -                       & -                          \\
              & AnyTeleop       & 1                  & -                    & -             & -                       & -                          \\
              & Open Teach      & 1                  & -                    & -             & -                       & -                          \\
\rowcolor[HTML]{EFEFEF} 
User 9        & Holo-Dex        & -                  & 0                    & -             & -                       & -                          \\
\rowcolor[HTML]{EFEFEF} 
              & AnyTeleop       & -                  & 0.4                  & -             & -                       & -                          \\
\rowcolor[HTML]{EFEFEF} 
              & Open Teach      & -                  & 0.8                  & 0             & -                       & 0.6                        \\
User 10       & Holo-Dex        & -                  & 0                    & -             & -                       & -                          \\
              & AnyTeleop       & -                  & 0.2                  & -             & -                       & -                          \\
              & Open Teach      & -                  & 0.6                  & 0.4           & 1                       & 1                          \\
\rowcolor[HTML]{EFEFEF} 
User 11        & Holo-Dex        & 1                  & -                    & -             & -                       & -                          \\
\rowcolor[HTML]{EFEFEF} 
              & AnyTeleop       & 1                  & -                    & -             & -                       & -                          \\
\rowcolor[HTML]{EFEFEF} 
              & Open Teach      & 1                  & -                    & -             & 0.8                     & 0.4                        \\
User 12        & Holo-Dex        & 1                  & -                    & -             & -                       & -                          \\
              & AnyTeleop       & 1                  & -                    & -             & -                       & -                          \\
              & Open Teach      & 1                  & -                    & -             & -                       & -                          \\
\rowcolor[HTML]{EFEFEF} 
User 13        & Holo-Dex        & 1                  & -                    & -             & -                       & -                          \\
\rowcolor[HTML]{EFEFEF} 
              & AnyTeleop       & 1                  & -                    & -             & -                       & -                          \\
\rowcolor[HTML]{EFEFEF} 
              & Open Teach      & 1                  & -                    & 0.6           & -                       & -                          \\
User 14       & Holo-Dex        & -                  & 0                    & -             & -                       & -                          \\
              & AnyTeleop       & -                  & 0.4                  & -             & -                       & -                          \\
              & Open Teach      & -                  & 0.6                  & -             & -                       & 0.8                        \\
\rowcolor[HTML]{EFEFEF} 
User 15        & Holo-Dex        & 1                  & -                    & -             & -                       & -                          \\
\rowcolor[HTML]{EFEFEF} 
              & AnyTeleop       & 1                  & -                    & -             & -                       & -                          \\
\rowcolor[HTML]{EFEFEF} 
              & Open Teach      & 1                  & -                    & 0.4           & -                       & -                          \\ \hline
\end{tabular}
\end{table*}

\begin{table*}[!th]
\centering
\caption{Average completion times for successful trials for the user study conducted across 15 individuals. Each user roughly performs 3 tasks on average. \textit{NS} denotes cases where no successes were achieved.}
\label{appendix:table:user_study_time}
\begin{tabular}{ccccccc}
\hline
\textbf{User} & \textbf{Method} & \multicolumn{5}{c}{\textbf{\begin{tabular}[c]{@{}c@{}}Average completion time for successful demonstrations (in s)\end{tabular}}} \\ \cline{3-7} 
\textbf{}     & \textbf{}       & \textbf{Flip Cube}     & \textbf{Pinch Grasp}     & \textbf{Pour}     & \textbf{Pick and Place}     & \textbf{Open Box of Mints}    \\ \hline
\rowcolor[HTML]{EFEFEF} 
User 1        & Holo-Dex        & 4.6                    & NS                       & -                 & -                           & -                             \\
\rowcolor[HTML]{EFEFEF} 
              & AnyTeleop       & 20.2                   & 22.5                     & -                 & -                           & -                             \\
\rowcolor[HTML]{EFEFEF} 
              & Open Teach      & 5.4                    & 18.6                     & 66                & -                           & -                             \\
User 2        & Holo-Dex        & -                      & 17.5                     & -                 & -                           & -                             \\
              & AnyTeleop       & -                      & 18.9                     & -                 & -                           & -                             \\
              & Open Teach      & -                      & 20.6                     & -                 & 29.7                        & 12.2                          \\
\rowcolor[HTML]{EFEFEF} 
User 3        & Holo-Dex        & 5.4                    & NS                       & -                 & -                           & -                             \\
\rowcolor[HTML]{EFEFEF} 
              & AnyTeleop       & 18.3                   & 7.8                      & -                 & -                           & -                             \\
\rowcolor[HTML]{EFEFEF} 
              & Open Teach      & 5.1                    & 12.6                     & -                 & -                           & 11.3                          \\
User 4        & Holo-Dex        & 11                     & NS                       & -                 & -                           & -                             \\
              & AnyTeleop       & 13.2                   & 31.4                     & -                 & -                           & -                             \\
              & Open Teach      & 6.2                    & 7.5                      & -                 & 16.9                        & 48.4                          \\ 
\rowcolor[HTML]{EFEFEF} 
User 5        & Holo-Dex        & -                      & NS                       & -                 & -                           & -                             \\
\rowcolor[HTML]{EFEFEF} 
              & AnyTeleop       & -                      & 11.4                     & -                 & -                           & -                             \\
\rowcolor[HTML]{EFEFEF} 
              & Open Teach      & -                      & 10.9                     & 41.6              & 12.4                        & -                             \\
User 6        & Holo-Dex        & -                      & NS                       & -                 & -                           & -                             \\
              & AnyTeleop       & -                      & 12.7                     & -                 & -                           & -                             \\
              & Open Teach      & -                      & 10.5                    & -                 & 23.57                       & -                             \\
\rowcolor[HTML]{EFEFEF} 
User 7        & Holo-Dex        & -                      & NS                       & -                 & -                           & -                             \\
\rowcolor[HTML]{EFEFEF} 
              & AnyTeleop       & -                      & NS                       & -                 & -                           & -                             \\
\rowcolor[HTML]{EFEFEF} 
              & Open Teach      & -                      & 19.1                     & 21.49             & 49                          & 37.8                          \\
User 8        & Holo-Dex        & 6.5                    & -                        & -                 & -                           & -                             \\
              & AnyTeleop       & 5.4                    & -                        & -                 & -                           & -                             \\
              & Open Teach      & 4.7                    & -                        & -                 & -                           & -                             \\
\rowcolor[HTML]{EFEFEF} 
User 9        & Holo-Dex        & -                      & NS                       & -                 & -                           & -                             \\
\rowcolor[HTML]{EFEFEF} 
              & AnyTeleop       & -                      & 49.9                     & -                 & -                           & -                             \\
\rowcolor[HTML]{EFEFEF} 
              & Open Teach      & -                      & 65.3                     & NS                & -                           & 32.21                         \\
User 10       & Holo-Dex        & -                      & NS                       & -                 & -                           & -                             \\
              & AnyTeleop       & -                      & 48                       & -                 & -                           & -                             \\
              & Open Teach      & -                      & 30.8                     & 40.3              & 48.7                        & 21.3                          \\
\rowcolor[HTML]{EFEFEF} 
User 11       & Holo-Dex        & 6.7                    & -                        & -                 & -                           & -                             \\
\rowcolor[HTML]{EFEFEF} 
              & AnyTeleop       & 11.5                   & -                        & -                 & -                           & -                             \\
\rowcolor[HTML]{EFEFEF} 
              & Open Teach      & 5.6                    & -                        & -                 & 21.8                        & 15.7                          \\
User 12       & Holo-Dex        & 6.2                    & -                        & -                 & -                           & -                             \\
              & AnyTeleop       & 11                     & -                        & -                 & -                           & -                             \\
              & Open Teach      & 3.8                    & -                        & -                 & -                           & -                             \\
\rowcolor[HTML]{EFEFEF} 
User 13       & Holo-Dex        & 8.9                    & -                        & -                 & -                           & -                             \\
\rowcolor[HTML]{EFEFEF} 
              & AnyTeleop       & 14.2                   & -                        & -                 & -                           & -                             \\
\rowcolor[HTML]{EFEFEF} 
              & Open Teach      & 5.8                    & -                        & 18.1              & -                           & -                             \\
User 14       & Holo-Dex        & -                      & NS                       & -                 & -                           & -                             \\
              & AnyTeleop       & -                      & 49.9                     & -                 & -                           & -                             \\
              & Open Teach      & -                      & 65.3                     & -                 & -                           & 132.5                         \\
\rowcolor[HTML]{EFEFEF} 
User 15       & Holo-Dex        & 13.2                   & -                        & -                 & -                           & -                             \\
\rowcolor[HTML]{EFEFEF} 
              & AnyTeleop       & 14.6                   & -                        & -                 & -                           & -                             \\
\rowcolor[HTML]{EFEFEF} 
              & Open Teach      & 6.3                    & -                        & 53.1              & -                           & -                             \\ \hline
\end{tabular}
\end{table*}


\end{document}